\documentclass[11pt,a4paper]{article}
\usepackage{natbib}
\usepackage{graphicx}
\usepackage{amsmath}
\usepackage{amsfonts}
\usepackage{amssymb}
\usepackage{makeidx}
\usepackage{setspace}
\usepackage{geometry}
\usepackage{color}
\usepackage{hyperref}
\usepackage{xcolor}
\usepackage{multirow}
\usepackage{lineno}
\hypersetup{colorlinks, linkcolor={red!50!black}, citecolor={blue!50!black}, urlcolor={blue!80!black}}
\usepackage{xr}
\makeatletter

\newcommand*{\addFileDependency}[1]{
\typeout{(#1)}
\@addtofilelist{#1}
\IfFileExists{#1}{}{\typeout{No file #1.}}
}\makeatother

\title{Learning high-level visual representations from a child's perspective without strong inductive biases}

\author{%
  A. Emin Orhan$^\delta$ \hspace{1cm} Brenden M. Lake$^{\delta,\psi}$ \\
  $^\delta$Center for Data Science, $^\psi$Department of Psychology \\
  New York University\\
  \texttt{\{eo41,\,brenden\}@nyu.edu}}
\date{}

\begin{document}

\maketitle
\begin{abstract}
Young children develop sophisticated internal models of the world based on their visual experience. Can such models be learned from a child's visual experience without strong inductive biases? To investigate this, we train state-of-the-art neural networks on a realistic proxy of a child's visual experience without any explicit supervision or domain-specific inductive biases. Specifically, we train both embedding models and generative models on 200 hours of headcam video from a single child collected over two years and comprehensively evaluate their performance in downstream tasks using various reference models as yardsticks. On average, the best embedding models perform at a respectable 70\% of a high-performance ImageNet-trained model, despite substantial differences in training data. They also learn broad semantic categories and object localization capabilities without explicit supervision, but they are less object-centric than models trained on all of ImageNet. Generative models trained with the same data successfully extrapolate simple properties of partially masked objects, like their rough outline, texture, color, or orientation, but struggle with finer object details. We replicate our experiments with two other children and find remarkably consistent results. Broadly useful high-level visual representations are thus robustly learnable from a representative sample of a child's visual experience without strong inductive biases.
\end{abstract}

Young children develop powerful internal models of the visual world. Their visual abilities for object categorization \citep{bomba1983,murphy2004}, segmentation \citep{kellman1983}, and physical prediction \citep{spelke1992} emerge well within the first year. By the time children are 4-5 years old, their object recognition capabilities are already mature enough that they can outperform highly capable computer vision models in challenging real-world visual object recognition tasks in head-to-head comparisons \citep{ayzenberg2020,huber2022}.

Is it possible to learn such powerful internal models of the world from a child's experience without strong, domain-specific inductive biases? Versions of this \textit{nature vs.~nurture} question have been debated for centuries \citep{locke1690,leibniz1704} and they continue to shape our understanding of intelligence. In the last couple of decades, some developmental psychologists hypothesized various innate inductive biases related to objects, agents, and space \citep{kellman1983,spelke1992,spelke1994}, as well as biases governing the categorization and labeling of objects \citep{markman1990,merriman1989}. Others, on the other hand, argued for the feasibility of building internal models of the world without such inductive biases, relying instead on the richness of the developing child's experience \citep{elman1996}.

Here, we approach this age-old \textit{nature vs.~nurture} question through a modern lens: we investigate what today's highly generic deep neural networks can learn from a representative sample of a child's egocentric visual experience. We train state-of-the-art self-supervised learning (SSL) algorithms on a large-scale, longitudinal, developmentally realistic dataset of headcam videos recorded from the perspective of individual children \citep{sullivan2022}. The dataset comprises hundreds of hours of longitudinal, natural videos recorded over 26 months of early development. Distinctive to our work, we train models on data from each individual child, simulating the child's learning problem as closely as possible. By using highly generic architectures and learning algorithms, we seek to understand what kinds of perceptual capabilities might be learnable from a child's visual experience without strong inductive biases.

We train both image embedding models that can be used in a variety of downstream visual recognition, segmentation, or detection tasks, and generative models that can be used to generate images and assign likelihoods to them. We quantitatively evaluate the capabilities of the trained models, compare their performance against a battery of reference models, and provide qualitative insights into the properties of the learned representations.

\subsection*{Models}
\begin{figure}
    \centering
    \includegraphics[width=1.0\textwidth]{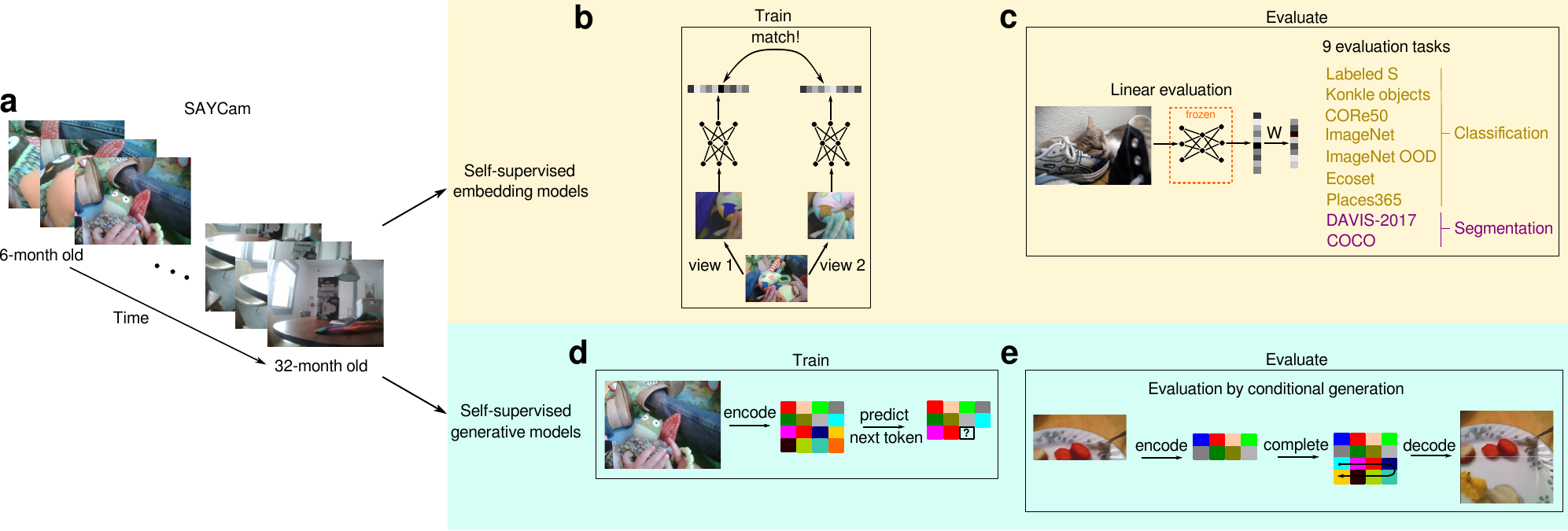}
    \caption{Schematic overview of the experiments. (\textbf{a}) Example video frames from longitudinal headcam recordings from one of the children in SAYCam \citep{sullivan2022}. (\textbf{b}) Training self-supervised embedding models. For purposes of illustration, a self-distillation type self-supervised learning algorithm is shown only, where the high-level goal is to learn representations that are similar across different views of the same image. (\textbf{c}) Evaluating the self-supervised embedding models. We evaluate the learned representations by training lightweight readouts on top of frozen features in 9 downstream classification or segmentation tasks. (\textbf{d}) Training self-supervised generative models. Frames are encoded into a spatially downsampled discrete code with the help of an optimized codebook. An autoregressive transformer model is trained to predict the next token in the discrete code. (\textbf{e}) Evaluating the self-supervised generative models. The top half of an evaluation image is given as context to the model. The model completes the bottom half of the image in the latent space and the model-completed latent code is decoded back to the image space for evaluation.}
    \label{overview_fig}
\end{figure}

We train two distinct types of models on a representative sample of a child's visual experience: embedding models and generative models. Embedding models aim to learn high-level visual features that are useful for a variety of downstream visual tasks. Generative models can generate novel images (both conditional on a given context and unconditionally) and assign likelihoods to images, providing a complementary tool for examining the acquired knowledge. Here, we briefly describe the algorithms, architectures, and training/evaluation methods for these models (Figure~\ref{overview_fig}). The Appendix provides additional details.

\subsubsection*{Embedding models}

\textbf{Self-supervised learning algorithms:} SSL algorithms seek to learn useful, high-level representations from a dataset without using any explicit supervision signals like semantic labels. Instead, they use augmented views of the training examples to generate self-supervision signals (Figure~\ref{overview_fig}b). We train embedding models with three different visual SSL algorithms: DINO \citep{caron2021}, Mugs \citep{zhou2022}, and masked autoencoders (MAE) \citep{he2022}.

\textbf{Model architectures:} Since our goal is to address a question of learnability with minimal inductive biases, we choose highly generic model architectures with minimal inductive biases. In particular, we focus mainly on vision transformer (ViT) models \citep{dosovitskiy2020}. We train models in three sizes: ViT-S, ViT-B, ViT-L (with 21M, 85M, 306M parameters, respectively), all with 16$\times$16 patches. With DINO, we further train ViT-B models with 14$\times$14 patches, as well as a convolutional ResNeXt-50 (32x4d) model \citep{xie2017} with 25M parameters. 

Both ViTs and the ResNeXt model incorporate two main inductive biases: hierarchical composition and translation invariance. These are very generic inductive biases quite different from the stronger, more domain-specific inductive biases about language, objects, agents, categories, or places, sometimes hypothesized by psychologists. The ResNeXt model incorporates a further spatial inductive bias with its convolutional filters. Our implementation of ViTs, on the other hand, uses learned position embeddings that are initialized randomly, therefore ViTs effectively start out with no spatial inductive biases.

\textbf{Training data:} Our main goal is to evaluate what can be learned from a representative sample of the visual experience of a developing child. To this end, we use the SAYCam dataset \citep{sullivan2022}: a large-scale, longitudinal dataset of natural headcam videos recorded from the perspective of three young children (S, A, and Y) between the ages of 6 to 31 months (Figure~\ref{overview_fig}a). The dataset contains 194 hours of video from S (6-30 months), 141 hours of video from A (8-31 months), and 137 hours of video from Y (7-24 months) for a total of 472 hours of video. Data from each child consist of a series of continuous, uninstructed headcam recordings, usually 1-2 hours of recording per week. These contain both inside and outside recording episodes. Videos are subsampled at 5 frames/second, for a total of 9 million frames across three children. We train models on data from each child individually as well as on the combined data (denoted as SAY below). Further details regarding the dataset can be found in \cite{sullivan2022}.

\textbf{Reference models:} To compare SAYCam-learned representations with representations learned from static photographic images, we train ViT-B/14 models (with DINO) on ImageNet \citep{russakovsky2015} and randomly sampled subsets of ImageNet (100\%, 10\%, and 1\% of the training set). To compare SAYCam-learned representations with representations learned from other video datasets, we train ViT-B/14 models (with DINO) on 200-hour long subsets of Kinetics-700 \citep{smaira2020} and Ego4D \citep{grauman2022} datasets (denoted as Kinetics-200h and Ego4D-200h below). Kinetics-700 consists of very short YouTube clips of people performing various actions, whereas Ego4D consists of long, continuous, egocentric headcam recordings from adults. We finally consider a randomly initialized, untrained reference model with the same architecture as the other reference models (ViT-B/14).

\textbf{Evaluation:} We use seven different classification tasks and two different semantic segmentation tasks for evaluation (see Figure~\ref{overview_fig}c for the full list). These include a classification task based on a labeled subset of the data from child S in SAYCam (\textit{Labeled S}), common object recognition (\textit{ImageNet}) and image segmentation (\textit{COCO}) benchmarks as well as a place classification task (\textit{Places365}). Using a wide range of evaluation tasks and datasets allows us to arrive at a more complete and robust picture of the overall quality of the learned visual representations. To evaluate visual representations learned exclusively through SSL, we use either completely non-parametric evaluation methods or methods that involve learning only a single layer of learnable parameters on top of frozen features (Figure~\ref{overview_fig}c). 

\subsubsection*{Generative models}
\textbf{Self-supervised learning algorithm:} We train generative transformer models on child headcam data, using a VQGAN-GPT architecture. We first learn a discrete codebook with a VQGAN \citep{esser2021}, and then encode each video frame as a spatial grid of integers from the codebook. These codes are then flattened and fed into a GPT model to learn a prior over the video frames. The GPT model is trained with the standard autoregressive language modeling objective \citep{radford2019}, \textit{i.e.} predicting the next token given all previous tokens in the flattened code (Figure~\ref{overview_fig}d).

\textbf{Evaluation:} We consider conditional generation tasks where we take evaluation images, give the upper half of each image as context, and ask the model to complete the bottom half of the image conditional on the upper half (Figure~\ref{overview_fig}e).

\subsection*{Results}
\subsubsection*{Embedding models}
\begin{figure}
    \centering
    \includegraphics[width=0.87\textwidth]{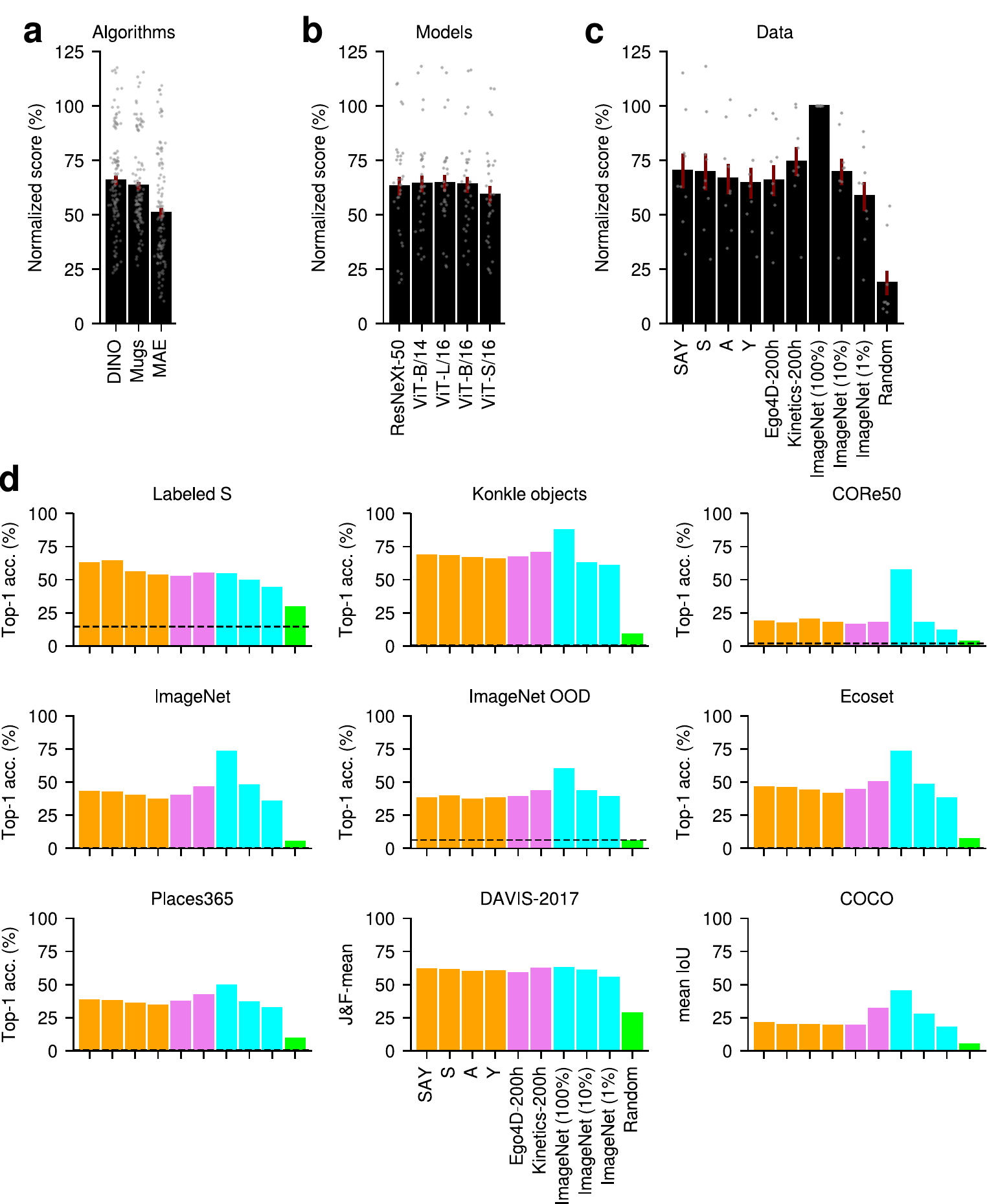}
    \caption{The effect of (\textbf{a}) algorithm, (\textbf{b}) model architecture, and (\textbf{c}) pretraining data on the performance in downstream evaluation tasks. All scores in \textbf{a}-\textbf{c} are relative to the ViT-B/14 model trained with DINO on all of ImageNet, our best model overall. Error bars represent standard errors. In \textbf{a}, means and standard errors are calculated over $n=108$ different \textit{model} (ViT-S/16, ViT-B/16, ViT-L/16) $\times$ \textit{data} (SAY, S, A, Y) $\times$ \textit{task} (9 evaluation tasks) combinations, represented by the individual gray dots. In \textbf{b}, the algorithm is fixed to DINO and the means and standard errors are calculated over $n=32$ different \textit{task} (all evaluation tasks except DAVIS-2017) $\times$ \textit{data} (SAY, S, A, Y) combinations. In \textbf{c}, the algorithm is fixed to DINO and the model architecture is fixed to ViT-B/14 and means and standard errors are calculated over $n=9$ evaluation tasks. (\textbf{d}) Performance of SAYCam-trained models compared with the reference models in all 9 evaluation tasks. As in \textbf{c}, here we again fix the algorithm to DINO and the model architecture to ViT-B/14. SAYCam-trained models are shown in {\color{orange}orange}; models trained on other video datasets are shown in {\color{magenta}magenta}; ImageNet trained models are shown in {\color{cyan}cyan}; the untrained reference model is shown in {\color{green}green}. Dashed horizontal lines show chance-level performance for the classification tasks. Note that performance is not normalized in \textbf{d}.}
    \label{embedding_summary_fig}
\end{figure}

\textbf{Quantitative summary:} Figure~\ref{embedding_summary_fig} summarizes the evaluation results of the embedding models, singling out the effects of SSL algorithm (Figure~\ref{embedding_summary_fig}a), model architecture (Figure~\ref{embedding_summary_fig}b), and pretraining data (Figure~\ref{embedding_summary_fig}c) on downstream task performance. In Figure~\ref{embedding_summary_fig}a-c, we normalize the performance on each task by the performance of a ViT-B/14 model trained with DINO on all of ImageNet, the overall best model. The DINO algorithm performs the best in our evaluations, with Mugs coming in second and MAE third. Different model architectures perform similarly except for ViT-S/16, which performs worse than the other models. Given these results, we focus most of our subsequent analyses on ViT-B/14 models trained with DINO, which is one of our best model$\times$algorithm combinations overall. 

Figure~\ref{embedding_summary_fig}c compares the performance of SAYCam-trained models against each of the reference models described above. Figure~\ref{embedding_summary_fig}d further splits Figure~\ref{embedding_summary_fig}c into different evaluation tasks. On average, SAYCam-trained models perform at 65-70\% of a model trained on the full ImageNet training set and they are generally comparable to a model trained with 10\% of ImageNet (SAY: $70.2\pm8.0$, S: $69.7\pm8.4$, A: $66.5\pm7.1$, Y: $64.5\pm7.2$, ImageNet-100\%: $100.0\pm0.0$, ImageNet-10\%: $69.7\pm6.0$)\footnote{Numbers represent means $\pm$ standard errors.}. Thus, although SAYCam-trained models are exposed to a very different type of data (less diverse, temporally extended, noisy headcam videos) than the ImageNet-trained model, they are able to recover a significant fraction of the ImageNet-trained model's performance.

All SAYCam-trained models substantially outperform the untrained reference model with random features (Random: $18.6\pm5.7$). Differences across individual children in SAYCam are relatively small (\textit{e.g.}~only 3\% relative difference between the approximately length-matched A and Y). Finally, the Ego4D-200h model performs comparably to the models trained on A and Y, and slightly worse than the model trained on the approximately length-matched S (Ego4D-200h: $65.6\pm7.1$), whereas the Kinetics-200h model performs better than all SAYCam-trained models (Kinetics-200h: $74.5\pm6.7$), although the difference is surprisingly small given the very different nature of the videos in Kinetics-200h compared with the videos in SAYCam or Ego4D (videos in Kinetics-200h are much shorter and more diverse in content).

\begin{figure}
    \centering
    \includegraphics[width=1.0\textwidth]{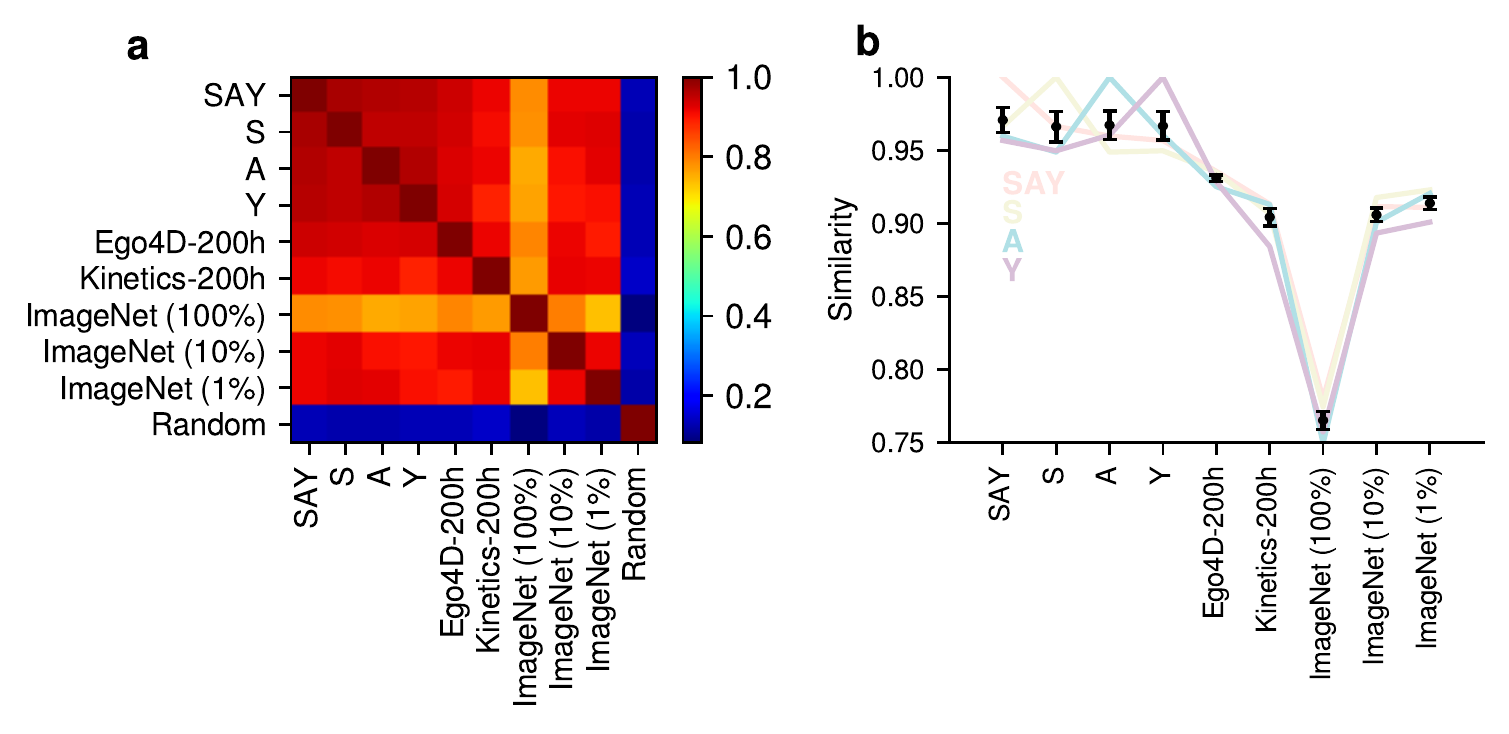}
    \caption{Representational similarity of ViT-B/14 models trained with DINO on different datasets. Representational similarity between a pair of models is measured as the symmetrized $R^2$ of the linear regression of the embeddings of one model with the embeddings of the other model, averaged over the embedding dimensions. We use the two-dimensional t-SNE embeddings of the 1000 ImageNet class means for this analysis. \textbf{a} shows the full similarity matrix, \textbf{b} singles out the models trained on SAYCam (SAY, S, A, and Y) and shows their representational similarities with each model except for the untrained Random model. Black dots and errors bars represent means and standard errors over the four SAYCam-trained models. The model trained on Ego4D-200h is the most representationally similar model to the SAYCam-trained models (Ego4D-200h: 93.1$\pm$0.2, Kinetics-200h: 90.4$\pm$0.6, ImageNet-10\%: 90.6$\pm$0.5, ImageNet-1\%: 91.4$\pm$0.2).}
    \label{model_similarity_fig}
\end{figure}

\begin{figure}
    \centering
    \includegraphics[width=1.0\textwidth]{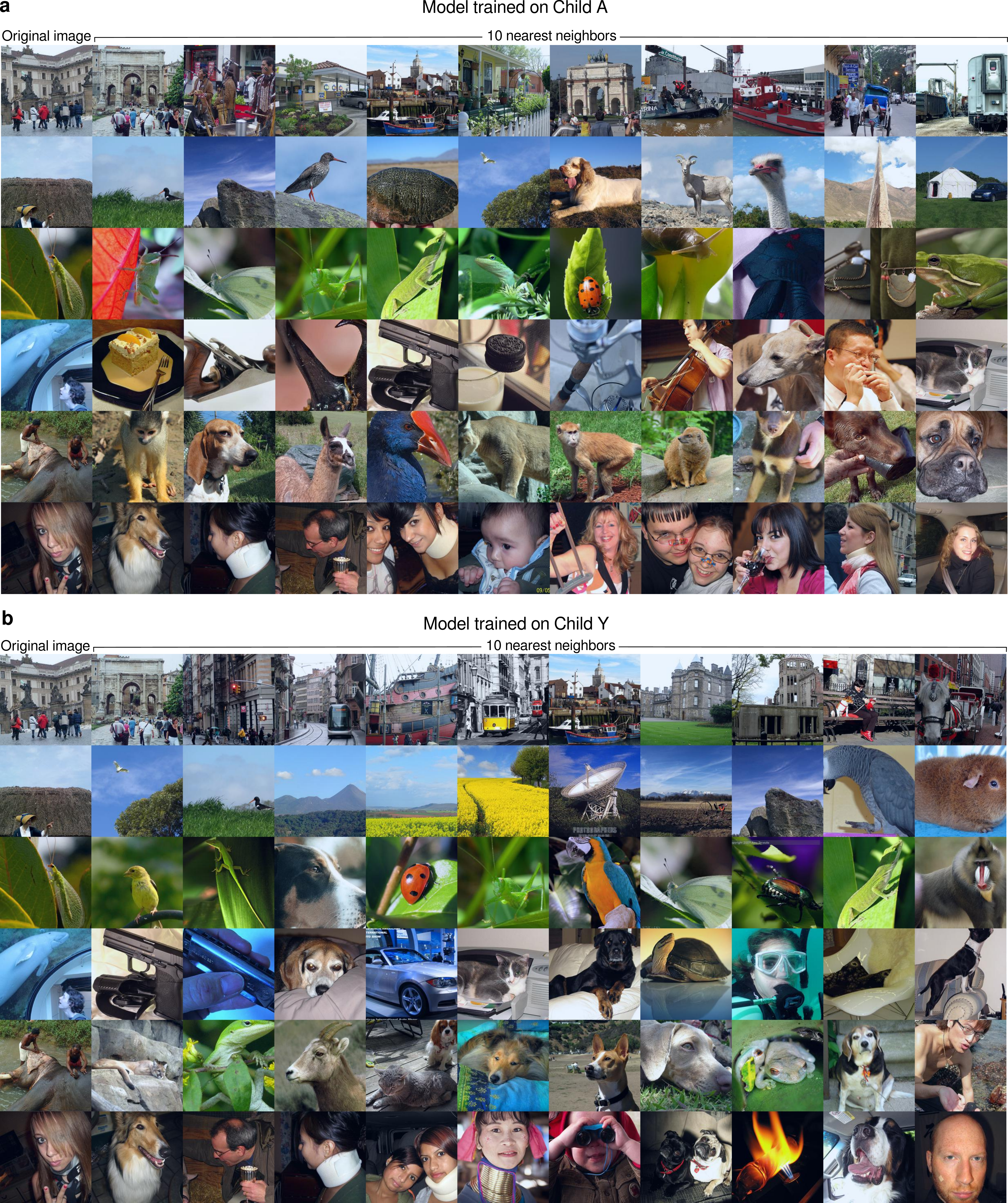}
    \caption{Nearest neighbors in the embedding space of ViT-B/14 models trained with DINO on the headcam data from child A (\textbf{a}) and from child Y (\textbf{b}).}
    \label{nearest_neighbors_ay_fig}
\end{figure}

The following qualitative analyses focus on models trained with the headcam data from child S only. The results for the other two children are qualitatively similar and they can be found in the Appendix (Figures~\ref{atts_a_fig}-\ref{tsne_y_fig} and \ref{model_similarity_fig}-\ref{nearest_neighbors_ay_fig}).

\begin{figure}
    \centering
    \includegraphics[width=0.88\textwidth]{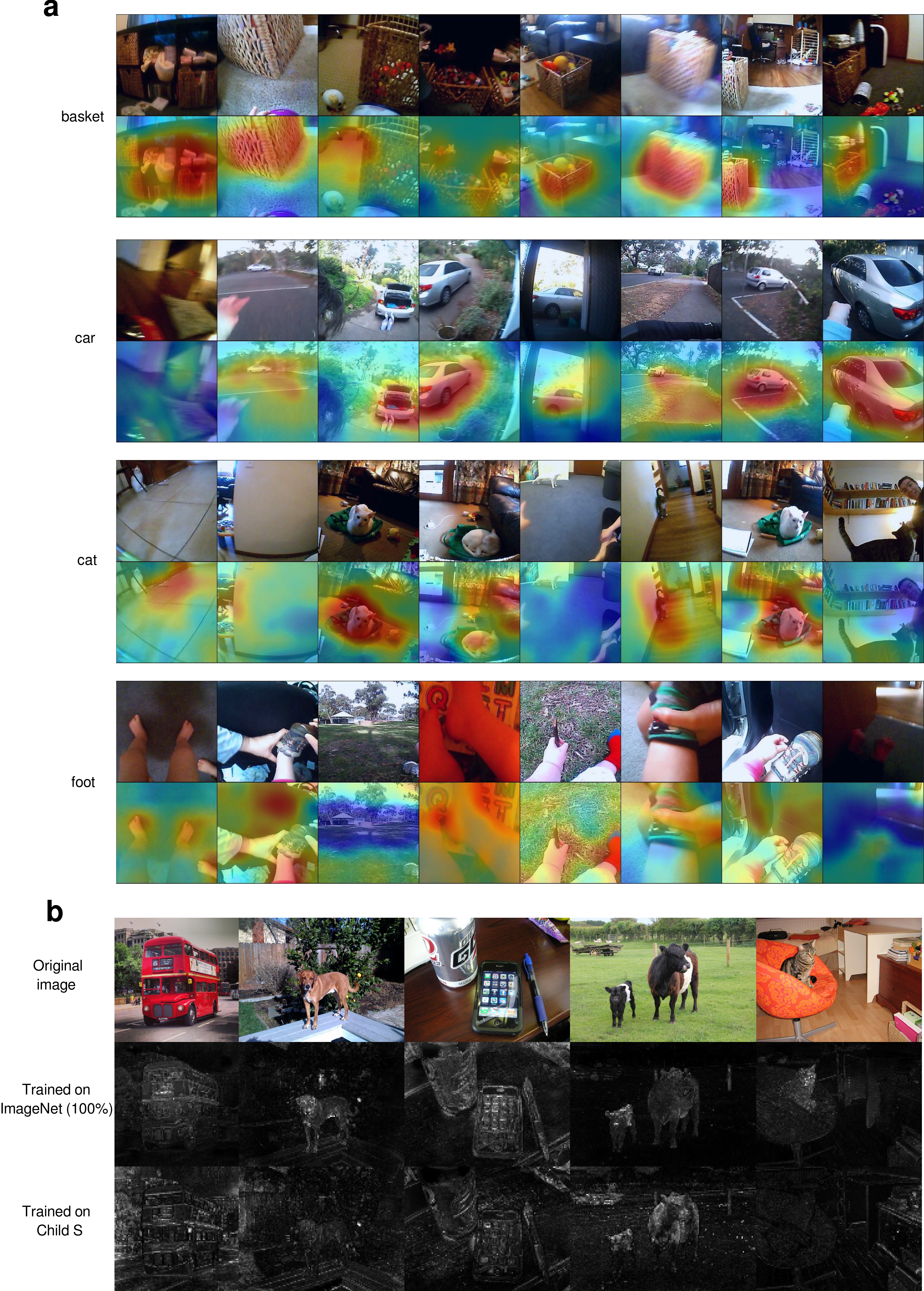}
    \caption{(\textbf{a}) Class activation maps (CAMs) for four different classes in \textit{Labeled S}: \textit{basket}, \textit{car}, \textit{cat}, \textit{foot}. In each case, the top row shows the original images, the bottom row shows the corresponding class activation maps. The class activation maps shown here are from a ResNeXt-50 model trained with DINO on data from child S only. More examples can be found at \href{https://github.com/eminorhan/dino/tree/master/cams}{this link}. (\textbf{b}) Original images from COCO and the corresponding attention maps (averaged over all attention heads) for ViT-B/14 models trained on all of ImageNet training set or on data from child S in SAYCam, respectively. The attention maps were computed with respect to the \texttt{cls} token.}
    \label{atts_cams_fig}
\end{figure}

\textbf{Learning to localize semantic categories without location supervision:} The semantic segmentation results in Figure~\ref{embedding_summary_fig}d (\textit{DAVIS-2017} and \textit{COCO}) show visual representations learned from a child's headcam data are much better than random representations at localizing semantic categories in an image, given dense (pixel-level) semantic feedback. These representations can also support localizing semantic categories without any explicit location feedback, using only information from a linear classifier trained on a downstream classification task. The last-layer feature maps of the model can be linearly combined with the classifier weights for a given class, generating a class activation map or CAM \citep{zhou2016}. Figure~\ref{atts_cams_fig}a illustrates CAMs for four different categories from the \textit{Labeled S} evaluation dataset. Qualitatively, the semantic localization obtained from CAMs is reasonably accurate in many, though not all, cases. Common failure cases include difficulties with localizing smaller objects and overbroad activation maps that extend into neighboring objects or surfaces. This may be related to the relatively global, background-sensitive nature of the representations learned by models trained with the child headcam data, as discussed next.

\begin{figure}
    \centering
    \includegraphics[width=1.0\textwidth]{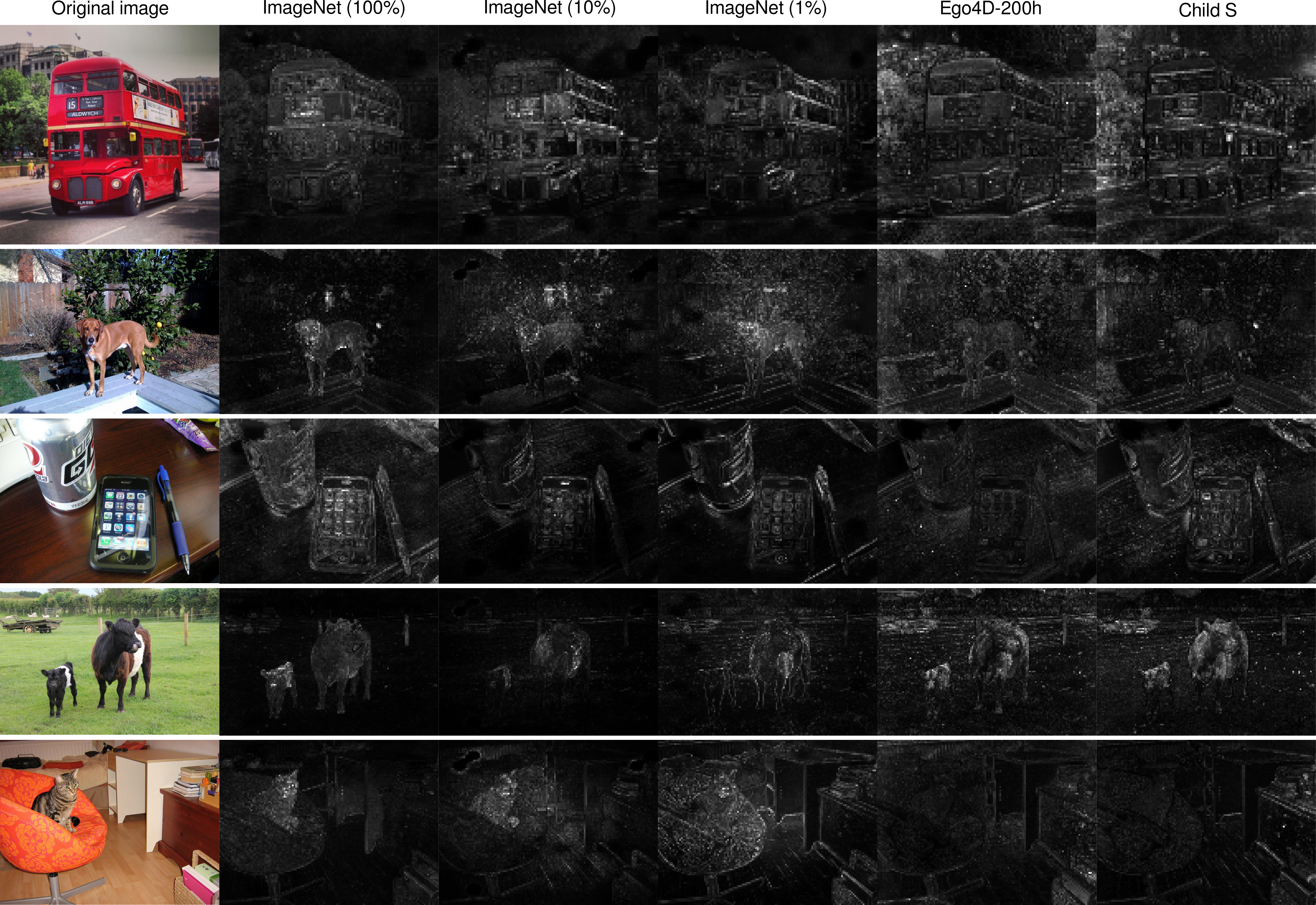}
    \caption{Attention maps (averaged over all attention heads) for ViT-B/14 models trained on different datasets: all of ImageNet (100\%), subsets of ImageNet (10\% and 1\%, respectively), Ego4D-200h, and Child S from SAYCam (from left to right). Models trained on effectively less data display signs of being more sensitive to background cues and low-level features like contours.}
    \label{atts_all_fig}
\end{figure}

\textbf{Learning more global, background-sensitive representations:} Visual representations learned from the child headcam data tend to be less object-centric and more sensitive to background and low-level surface features (\textit{e.g.}~contours) compared to ImageNet-learned representations. This is illustrated in Figure~\ref{atts_cams_fig}b, which compares the mean attention maps (averaged over all attention heads) of ViT-B/14 models trained on ImageNet and on the headcam data from child S. These observations are quantitatively supported by the performance of the models on CORe50 (Figure~\ref{embedding_summary_fig}d), which evaluates the background-invariance of the models' object representations. Models trained with small subsets of ImageNet are also less object-centric (Figure~\ref{atts_all_fig}), suggesting that learning object-centric, background-invariant representations may require seeing the foreground objects against a sufficiently large and diverse set of backgrounds.

\begin{figure}
    \centering
    \includegraphics[width=1.0\textwidth]{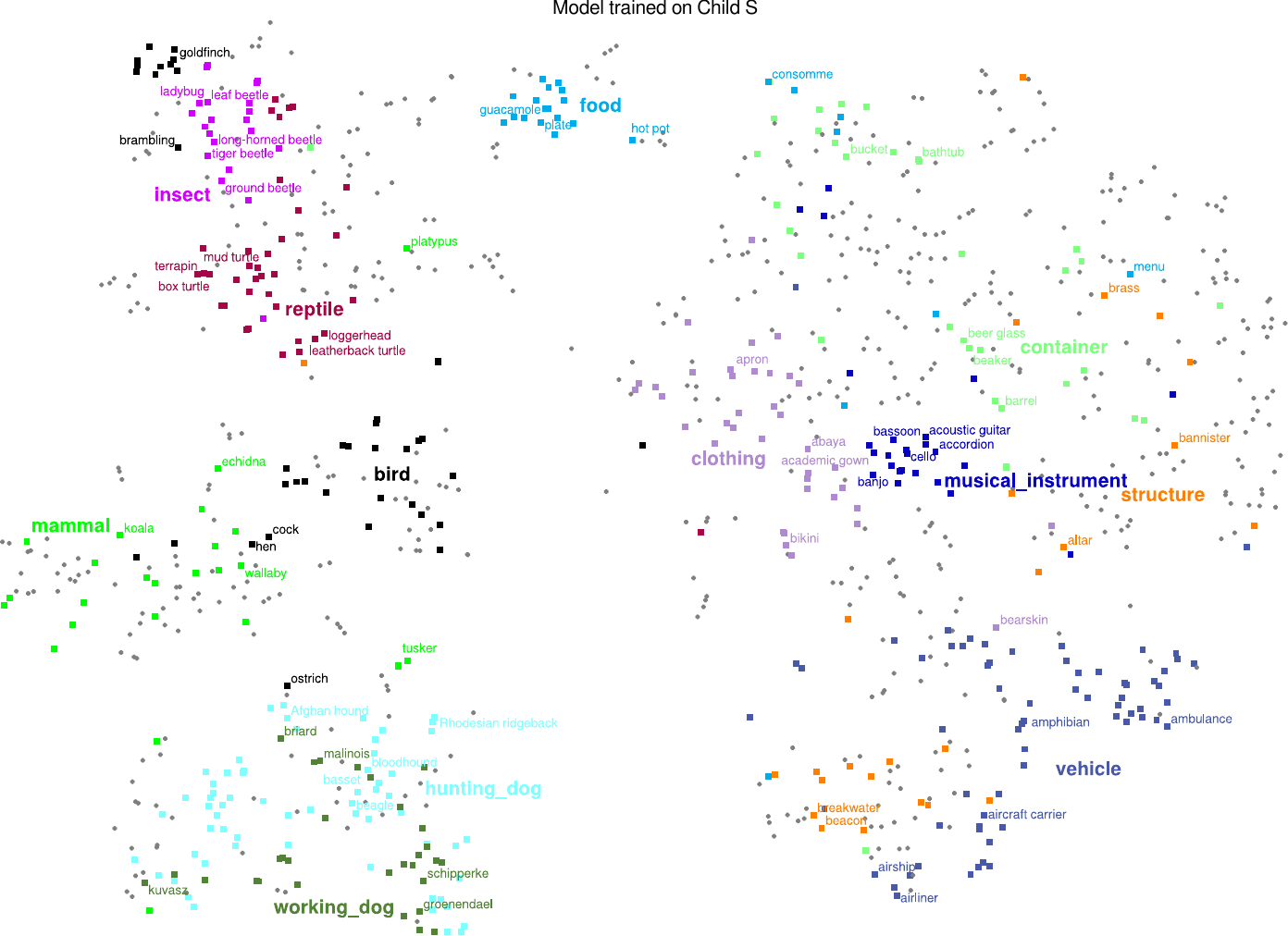}
    \caption{t-SNE visualization of the 1000 ImageNet classes for a ViT-B/14 DINO model trained on data from child S only. Each point corresponds to a different ImageNet class. The class embeddings are computed as the mean embedding over all validation images belonging to that class.~Different colors represent 12 different super-classes (indicated in larger font) extracted from the WordNet hierarchy. Five classes are labeled individually for each super-class. The other classes are not labeled individually for legibility. The visualizations for models trained on the other childrens' data are qualitatively very similar (Figures~\ref{tsne_a_fig}-\ref{tsne_y_fig}). More t-SNE visualizations can be found at \href{https://github.com/eminorhan/dino/tree/master/tsnes}{this link}.}
    \label{tsne_fig}
\end{figure}

\begin{figure}
    \centering
    \includegraphics[width=1.0\textwidth]{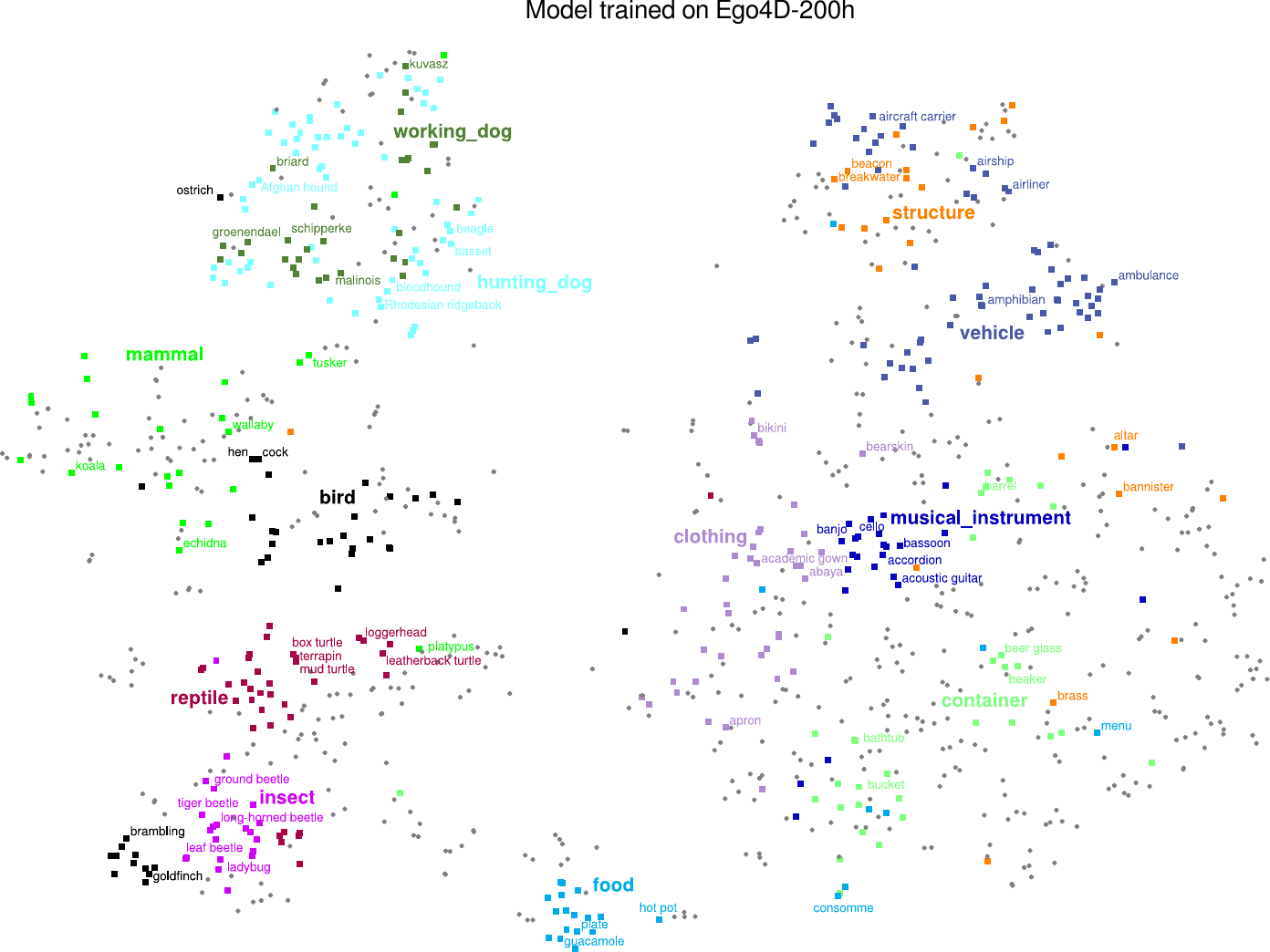}
    \caption{t-SNE visualization of the 1000 ImageNet classes for a ViT-B/14 model trained with DINO on a 200-hour subset of Ego4D (Ego4D-200h).}
    \label{tsne_ego_fig}
\end{figure}

\begin{figure}
    \centering
    \includegraphics[width=1.0\textwidth]{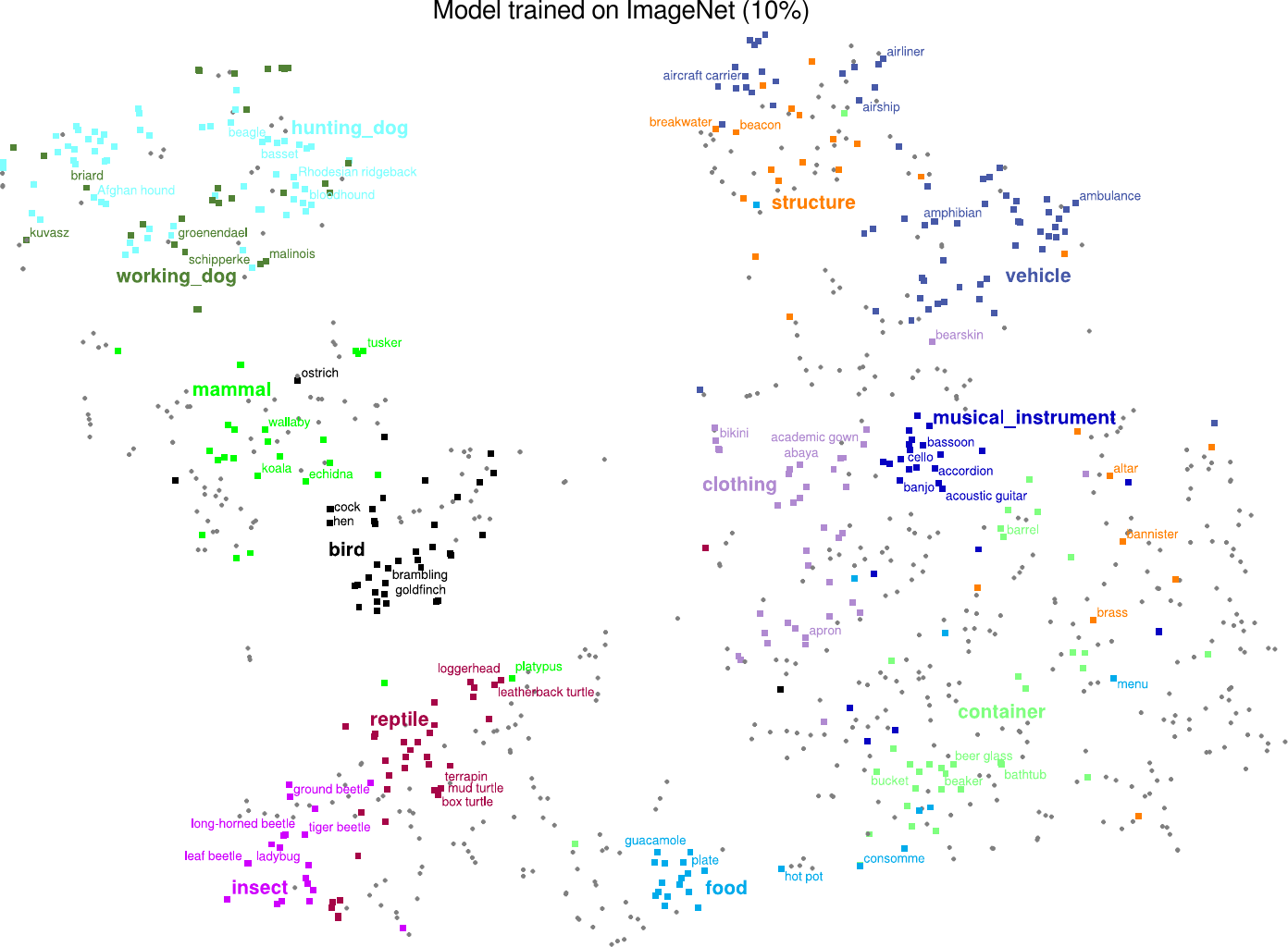}
    \caption{t-SNE visualization of the 1000 ImageNet classes for a ViT-B/14 model trained with DINO on 10\% of ImageNet.}
    \label{tsne_imagenet10_fig}
\end{figure}

\begin{figure}
    \centering
    \includegraphics[width=1.0\textwidth]{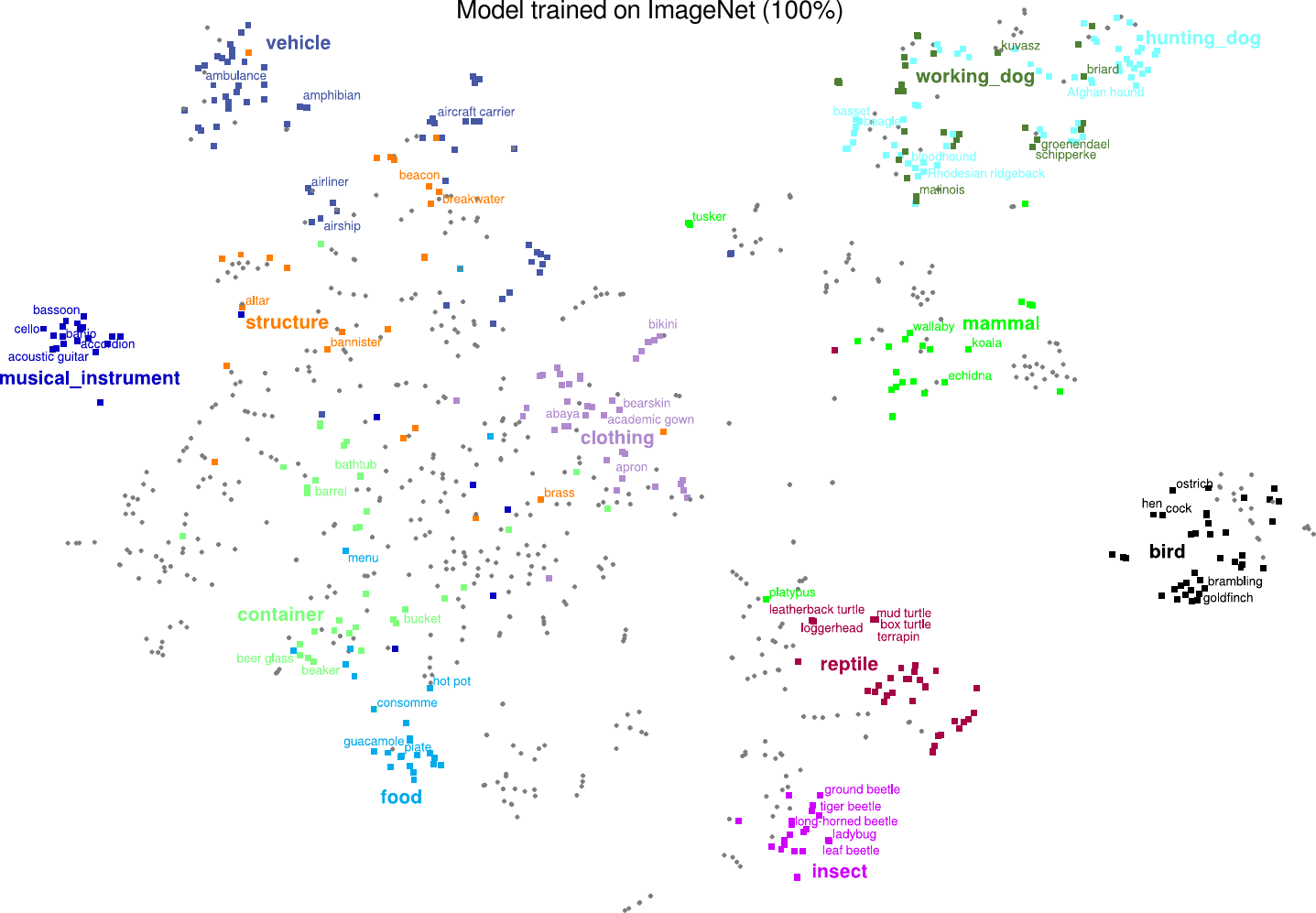}
    \caption{t-SNE visualization of the 1000 ImageNet classes for a ViT-B/14 model trained with DINO on all of ImageNet.}
    \label{tsne_imagenet_fig}
\end{figure}

\begin{figure}
    \centering
    \includegraphics[width=1.0\textwidth]{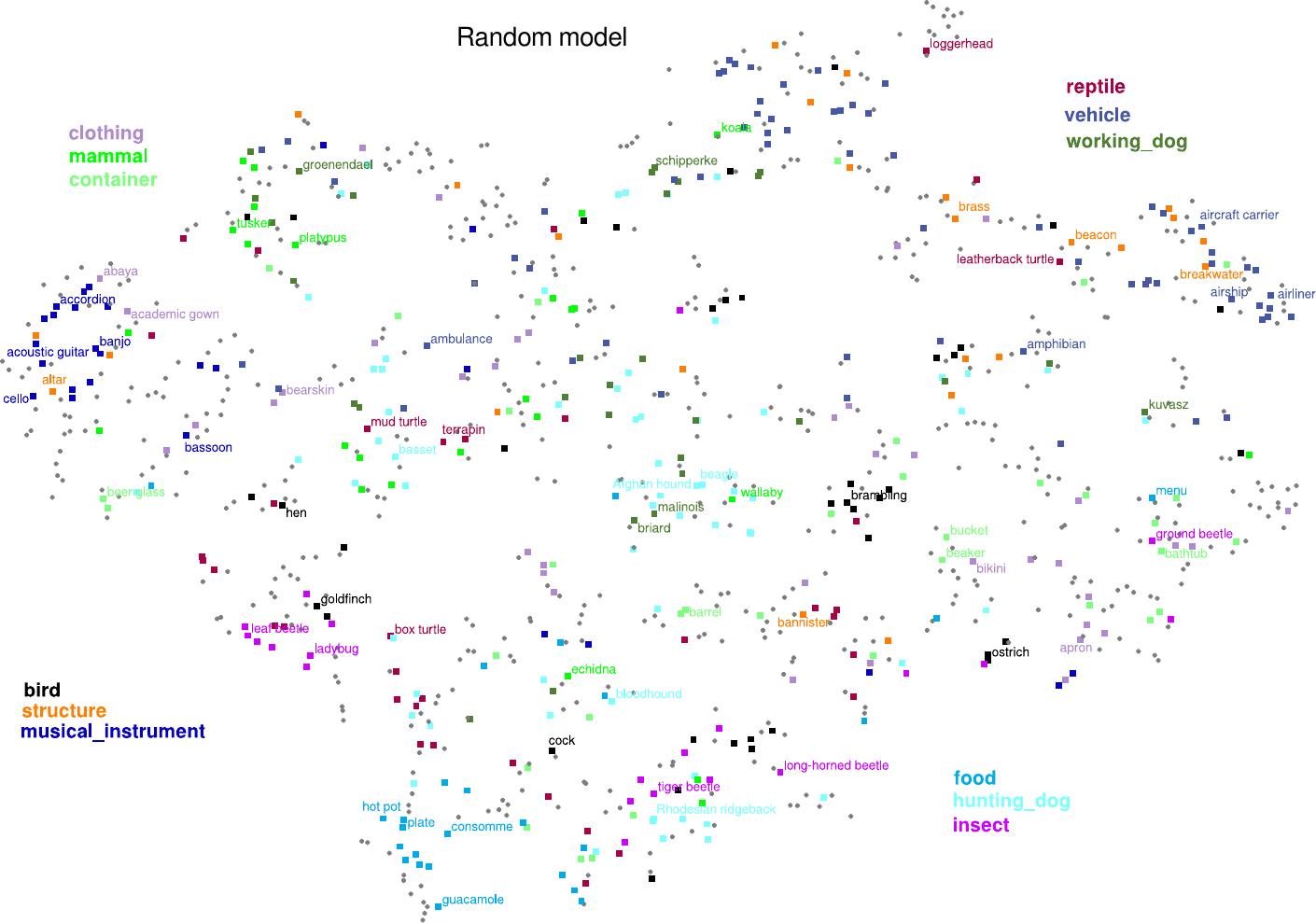}
    \caption{t-SNE visualization of the 1000 ImageNet classes for an untrained, randomly initialized ViT-B/14 model. Compared to the SAYCam-trained models, the embeddings display much weaker semantic structure.}
    \label{tsne_random_fig}
\end{figure}

\textbf{Learning broad semantic categories without any labeled examples:} A rich semantic structure emerges in the embedding space of the models trained with the child headcam data. Figure~\ref{tsne_fig} shows a t-SNE visualization \citep{vandermaaten2008} of the mean embeddings of the 1000 ImageNet classes (estimated over the validation set) obtained from a model trained on child S. Classes belonging to the same broad semantic categories such as \textit{dogs, birds, reptiles, insects, vehicles, musical instruments, food, clothing, etc.}~tend to be clustered together in the embedding space. Notably, the model learns this structure automatically without any labeled examples. This structure is either absent or much weaker in the embedding space of untrained, random models (Figure~\ref{tsne_random_fig}; also see Figures~\ref{tsne_a_fig}-\ref{tsne_imagenet_fig} for embeddings from other trained models).~Interestingly, the semantic structure that emerges in the embedding spaces of SAYCam-trained models is representationally most similar to the semantic structure in a model trained with the egocentric headcam data from adults (Ego4D-200h), followed by the other models that perform similarly in the downstream evaluation tasks (Figure~\ref{model_similarity_fig}).

\begin{figure}
    \centering
    \includegraphics[width=1.0\textwidth]{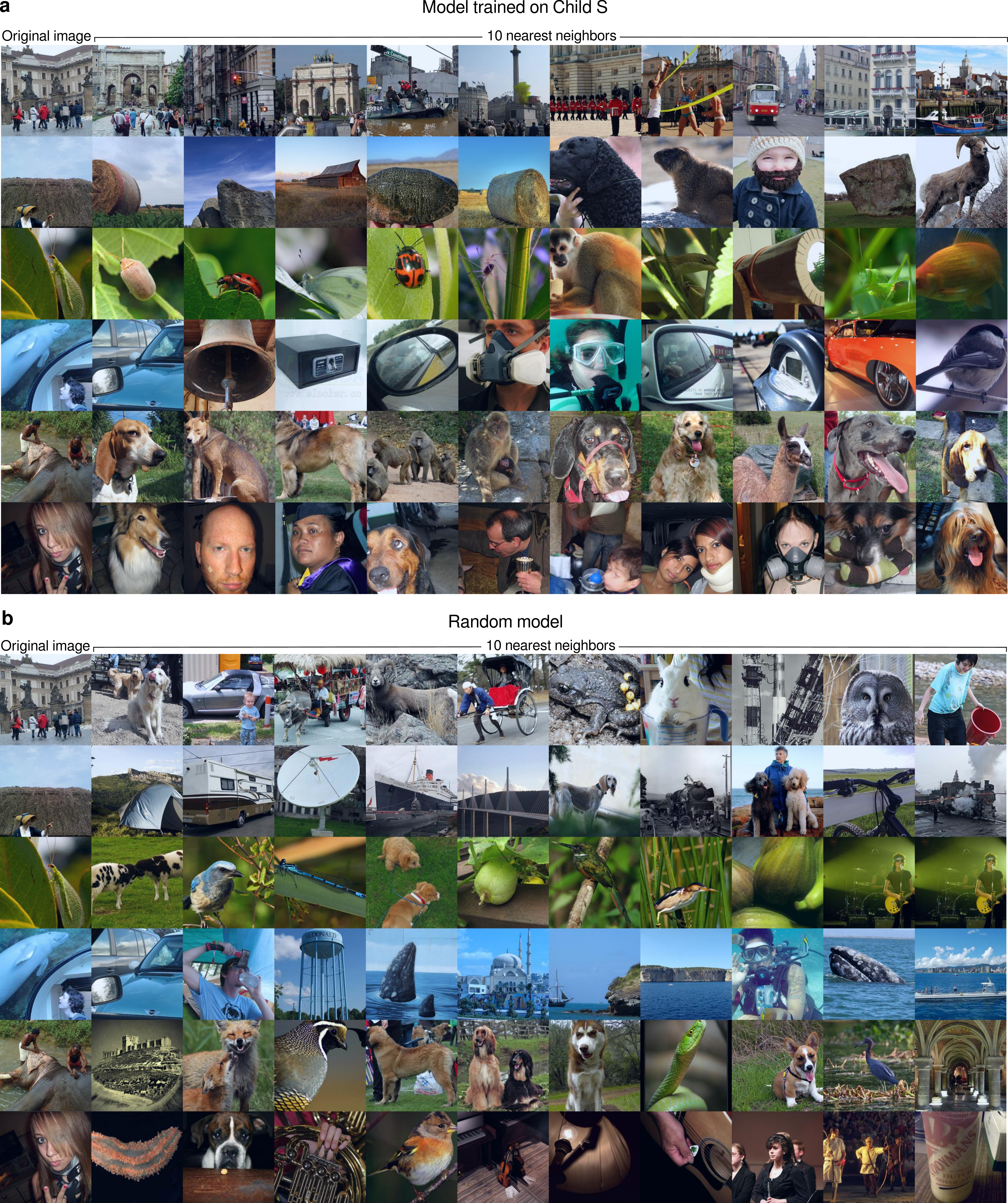}
    \caption{Nearest neighbors in the embedding space of a ViT-B/14 DINO model trained on child S (\textbf{a}) and in the embedding space of a random, untrained model with the same architecture (\textbf{b}). The leftmost column shows six query images, the next ten images in each row are the 10 nearest neighbors in the embedding space. Both the query and the nearest neighbors are from the ImageNet validation set. Nearest neighbors are with respect to the Euclidean metric.}
    \label{nearest_neighbors_fig}
\end{figure}

\begin{figure}
    \centering
    \includegraphics[width=1.0\textwidth]{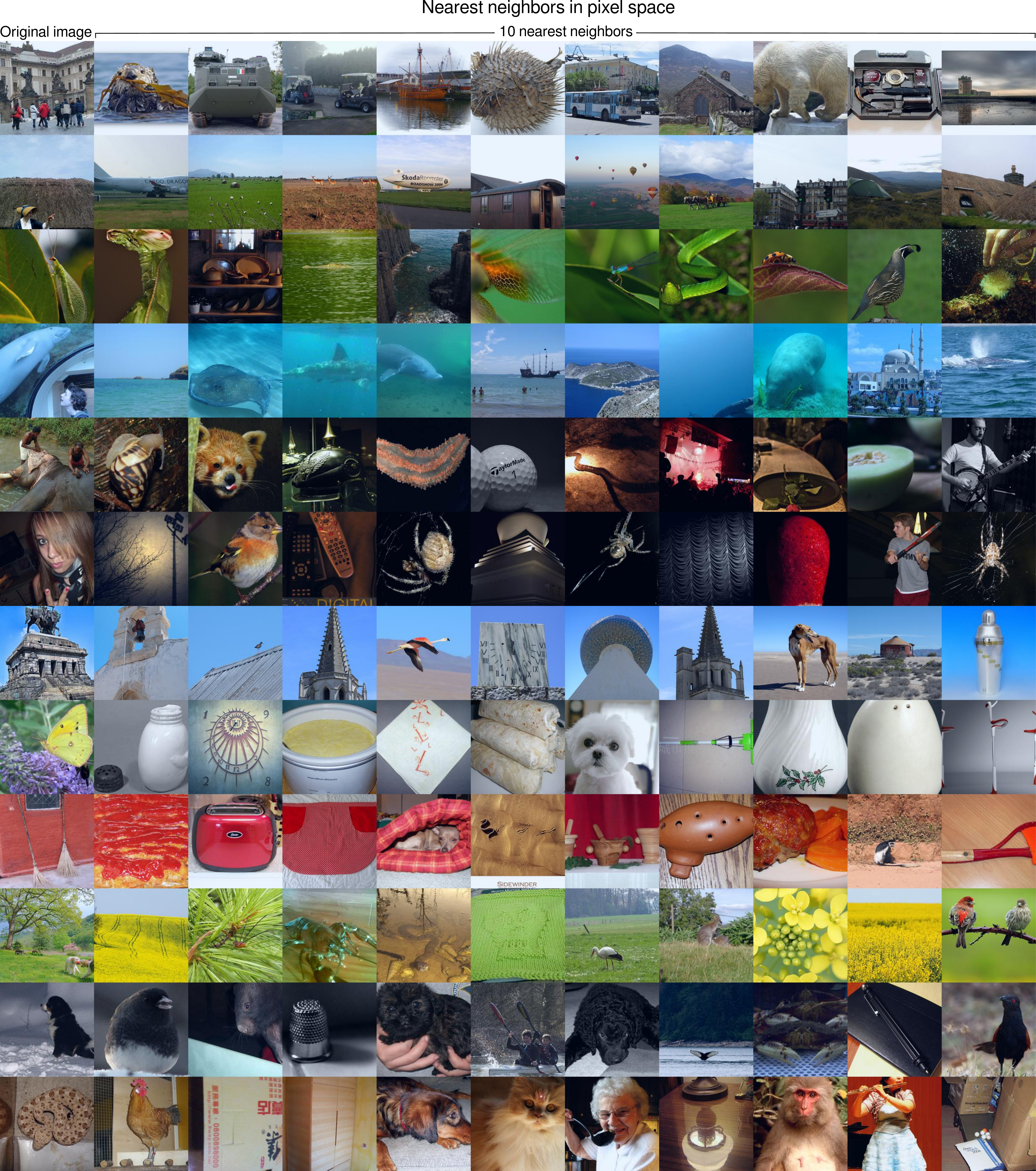}
    \caption{Nearest neighbors in the embedding space of pixels. For computational tractability, images are first downsampled so that the minor edge is 64 pixels long, then a 56$\times$56 central crop is taken to obtain the embedding.}
    \label{nearest_neighbors_pixels_fig}
\end{figure}

\textbf{Nearest neighbors reveal semantic structure in the embedding space:} Figure~\ref{nearest_neighbors_fig} shows query images from the ImageNet validation set (leftmost column) and their 10 nearest neighbors in two different embedding spaces. Retrievals from the embedding space of a model trained with the headcam data from child S are often semantically related to the query image (Figure~\ref{nearest_neighbors_fig}a). The failure cases usually preserve some semantic relationships (\textit{e.g.}~retrievals of dogs, primates, or other mammals for the elephant query in the fifth row of Figure~\ref{nearest_neighbors_fig}a) or display visual similarities with the texture or the overall shape of the object depicted in the query image (\textit{e.g.}~the thatched roof queried in the second row of Figure~\ref{nearest_neighbors_fig}a and the hay rolls retrieved in response to it have similar visual textures). The retrievals from the embedding space of an untrained, random model, on the other hand, seem to be primarily driven by the overall color similarity between the query and the retrieved item (Figure~\ref{nearest_neighbors_fig}b). A similar color-based similarity structure emerges in the embedding space of pixels as well (Figure~\ref{nearest_neighbors_pixels_fig}).

\subsubsection*{Generative models}
\begin{figure}
    \centering
    \includegraphics[width=1.0\textwidth]{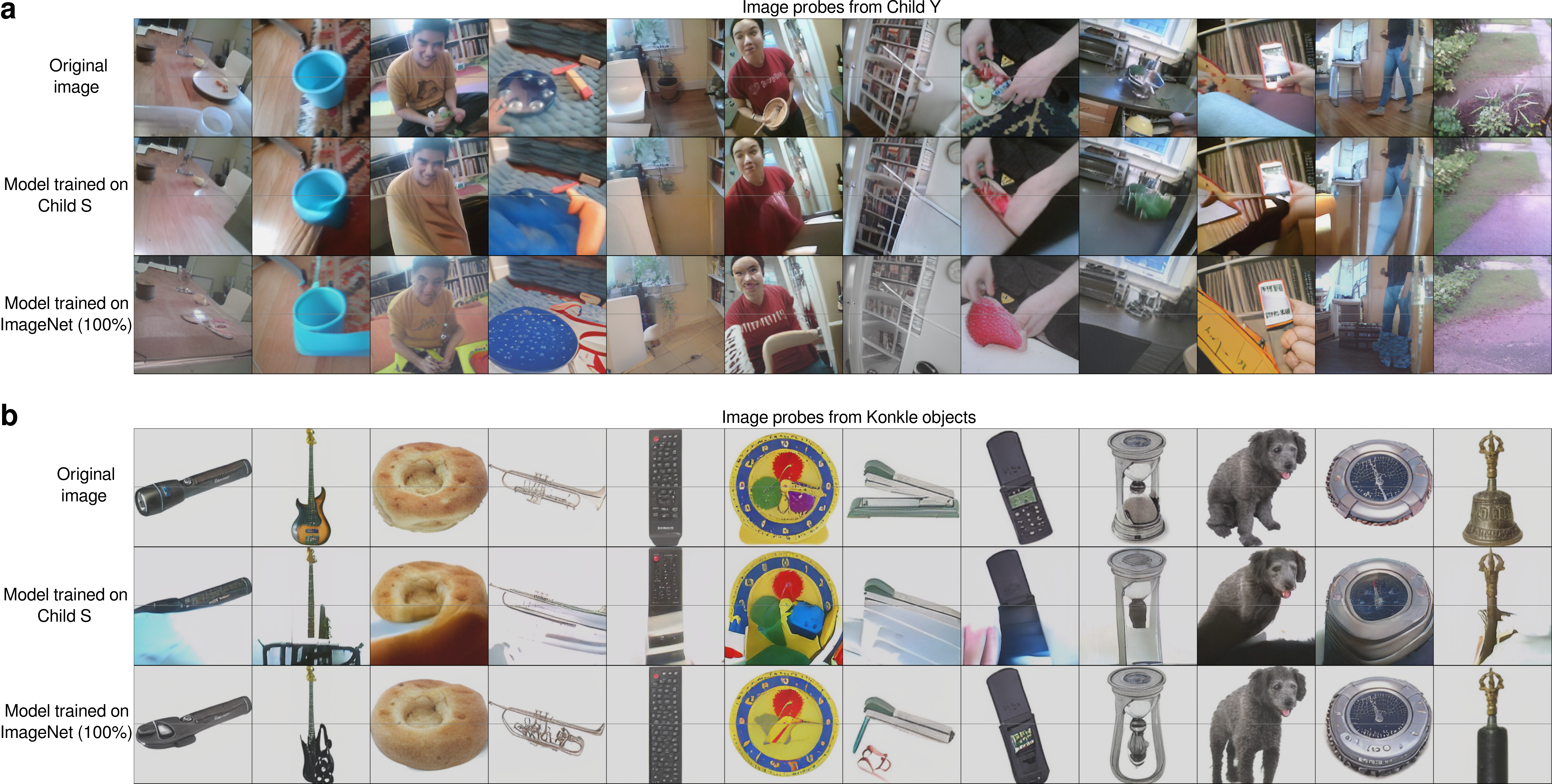}
    \caption{Conditional samples from two different models (trained on child S in SAYCam or on all of ImageNet) seeded with (\textbf{a}) images from child Y in SAYCam or with (\textbf{b}) images from the \textit{Konkle objects} dataset. In each case, the upper half of the image is given to the model as context, the lower half (separated by a line from the upper half) is generated by the model. All model completions are zero-shot (the models have not seen any prior examples from these datasets). More examples can be found at \href{https://github.com/eminorhan/silicon-menagerie/tree/master/gpt_samples}{this link}.}
    \label{vqgangpt_completion_fig}
\end{figure}

Generative models offer an alternative and intuitive route to studying learnability from a child's visual experience, as their outputs can be visualized directly. Here, we use an image completion task to probe the visual knowledge acquired by generative models trained on the child headcam data. We provide the model with the upper half of an image and generate the bottom half from the model with sampling. Figure~\ref{vqgangpt_completion_fig}a shows different images (columns) from child Y's data together with completions generated by a model trained on another child (child S) as well as a model trained on all of ImageNet. Similarly, Figure~\ref{vqgangpt_completion_fig}b shows different images from the \textit{Konkle objects} dataset and the corresponding completions. All of these completions are ``zero-shot'' in that the models have not seen any examples from these datasets during training. Although the model trained on child S can usually generate completions that match the color, texture, orientation, and rough outline of the object (or objects) given in the context (\textit{e.g.}~the compass in Figure~\ref{vqgangpt_completion_fig}b; second image from the right), it is not very successful at generating finer details of the objects (\textit{e.g.}~it is not very good at generating plausible looking legs for the dog in Figure~\ref{vqgangpt_completion_fig}b). The model trained on all of ImageNet, on the other hand, is much better at generating finer object details. We measure the quality of the completions generated by different models through Fr\'echet Inception distance (FID) scores evaluated on two datasets under different conditions (Table~\ref{fid_scores_table}). The FID scores broadly confirm our qualitative observations. In particular, the model trained on all of ImageNet consistently outperforms the SAYCam-trained models on images from the \textit{Konkle objects} dataset, although the generation quality of SAYCam-trained models on this dataset can be improved with a small amount of finetuning.

\section*{Discussion}
In this article, we investigated what state-of-the-art SSL algorithms can learn from a representative sample of a child's longitudinal, egocentric visual experience without strong inductive biases. Our analyses reveal both strengths and weaknesses of the representations learned from a child's visual experience with current SSL algorithms. On the one hand, with the equivalent of a few weeks of visual experience only, models trained with data from individual children already perform at 65-70\% of a high-performance ImageNet-trained model in a diverse range of downstream evaluation tasks (Figure~\ref{embedding_summary_fig}). They can also learn to localize semantic categories in an image without any explicit location supervision (Figure~\ref{atts_cams_fig}a), and they can learn broad semantic categories in an unsupervised way (Figure~\ref{tsne_fig}). Thus, despite significant differences between the visual experience of a developing child and the standard datasets used for training state-of-the-art computer vision models \citep{smith2017}, models trained with a realistic proxy of a child's visual experience still display highly non-trivial visual capabilities. These capabilities are also surprisingly consistent across models trained on different children in SAYCam (Figure~\ref{embedding_summary_fig}c; also see Figure \ref{model_similarity_fig}), even with substantial individual differences in the environments and behaviors of these children \citep{sullivan2022}. On the other hand, these models seem to be less object-centric than models trained with large-scale, photographic image datasets like ImageNet (Figure~\ref{atts_cams_fig}b) and in generative tests with out-of-domain stimuli, they seem to struggle with fine object details, even though they can successfully extrapolate the texture, color, orientation, and rough outlines of objects (Figure~\ref{vqgangpt_completion_fig}).

In our experiments, we used reference models trained on different types of visual data to better situate the capabilities of the SAYCam-trained models. Some of these reference models display visual capabilities comparable to the models trained on individual children in SAYCam (Figure~\ref{embedding_summary_fig}c):~\textit{e.g.}~ImageNet (10\%), Ego4D-200h, even Kinetics-200h to some extent, despite notable qualitative differences between these visual data. This result suggests a considerable degree of robustness in the emergence of these general visual capabilities. Some earlier works, on the other hand, emphasized the special properties of child-centric visual data from a representation learning perspective \citep{smith2017,bambach2018,zaadnoordijk2022}. Our results are not necessarily inconsistent with these studies: because we focused on relatively broad measures of performance in our qualitative and quantitative evaluations, we cannot rule out more fine-grained differences between the models that might be hidden behind their comparable overall performance. However, isolating the causes of such potential fine-grained differences would be difficult in our case, as our reference datasets differ across many dimensions.

What are the implications of our results for the \textit{nature vs.~nurture} question regarding the acquisition of basic visual capabilities such as real-world object recognition? Motivated by the early emergence of some visual capabilities in infants, developmental psychologists postulated various innate constraints related to objects, agents, space, and categories \citep{kellman1983,spelke1992,spelke1994,markman1990,merriman1989}, hypothesized to be critical for subsequent learning. However, a rigorous computational test of these claims requires considering both a sufficiently realistic proxy of a child's actual visual experience, and powerful, generic, scalable learning algorithms and models. Arguably for the first time in history, we now have both ingredients thanks to advances in the collection of large-scale longitudinal developmental datasets like SAYCam \citep{sullivan2022}, and advances in deep learning, giving us powerful generic learning algorithms and architectures. Together with a handful of other recent studies \citep{bambach2018,orhan2020,lee2021,zhuang2021,zhuang2022}, this work is among the first to take advantage of these new opportunities to address fundamental questions in cognitive science. Our results, for example, suggest that strong inductive biases like a taxonomic generalization bias or an innate ability to segment objects may be unnecessary, as our generic self-supervised models already do a reasonably good job of learning to segment objects in images (Figure~\ref{atts_cams_fig}a) or to categorize objects based on their kind (Figure~\ref{tsne_fig}) from limited and noisy visual data available to a child without such inductive biases. However, the models' ability to cleanly segment and generate objects is imperfect (Figures~\ref{atts_cams_fig}b and \ref{vqgangpt_completion_fig}), so it remains an open empirical question if they can attain human-level understanding of objects by simply being trained on developmentally more realistic amounts of data or if stronger object-centric inductive biases may still be necessary to achieve this \citep{kellman1983,spelke1992,locatello2020}.

There are a number of differences between our experimental setting and the actual learning problem faced by children. These differences should be kept in mind when considering the implications of our results for developmental psychology. First, even the combined data from SAYCam amount to roughly 40 days of visual experience (factoring in 12 hours of sleep per day). To extend this to developmentally realistic amounts of data would require roughly two orders of magnitude more data than we currently have. The capabilities of the current models would undoubtedly improve with additional data at this scale even without any other changes, but it is an open empirical question how much they would improve. Second, here we only considered visual data, but a child's actual experience is multimodal, with auditory, haptic, and sensorimotor components in addition to vision. The capabilities of the current models would again likely improve with these complementary sources of information. Third, our models are trained with stochastic gradient descent, which is biologically implausible in the context of deep networks \citep{lillicrap2020}. To the extent that biological learning must satisfy demanding constraints that are not relevant for deep learning, our results may overestimate what can be learned from a child's visual experience with biologically plausible learning mechanisms. Compared to deep learning models, this may necessitate more reliance on innate inductive biases in humans.

Another difference is that children are interactive learners. They learn their own behavioral policies regarding how to interact with objects or other agents in the environment. This allows them to shape and structure their own sensory experiences. Our models, on the other hand, are passive learners. The learnability results here thus relate to what is learnable from a visual stream that is, to some extent, already structured by the child. Interactive models that can actively shape their own experiences, like children do, might learn more effectively compared to passive learners \citep{gureckis2012}, in which case our results would underestimate what can be learned from child-like visual experience without strong inductive biases.

There are also important differences between the raw visual inputs received by our models \textit{vs.}~children. The SAYCam frames have relatively low spatial resolution (640$\times$480 pixels) compared to the human retina. They contain a significant amount of motion blur artifacts and the image quality is generally poor in low lighting conditions. Efforts to collect higher quality headcam data with better cameras are already under way \citep{long2022}. Modern SSL algorithms often use heavy data augmentation strategies like color jittering or random resized cropping (the particular data augmentations used by each of our SSL algorithms are detailed in the Appendix). These augmentations increase the effective sample size to the benefit of the models. It is unclear whether similar processes in humans could implicitly expand the input in a biologically plausible way. Foveation represents an interesting example in this respect with its functional similarity to random resized cropping.

We hope that our work will inspire new collaborations between machine learning and developmental psychology \citep{smith2017,zaadnoordijk2022,moore2022}, as the impact of modern deep learning on developmental psychology has been relatively limited thus far. One key reason for this is the data gap between machine and human learners: for example, today's computer vision models are typically trained with visual data that are very different in content, style, and amount from a child's visual experience (millions, sometimes billions, of static photographic pictures scraped from the internet \textit{vs.}~a few years of continuous, egocentric data streams from the world). Here, we bridged this data gap by training the same models on a realistic proxy of a child's egocentric visual experience and demonstrating these models' powerful visual capabilities. Future algorithmic advances, combined with richer and larger developmental datasets, can be evaluated through the same approach, further enriching our understanding of what can be learned from a child's experience with minimal inductive biases.

\section*{Acknowledgments} 
We thank Wai Keen Vong, Alexa Tartaglini, and Mengye Ren for helpful discussions and comments on an earlier version of this manuscript. This work was supported by the DARPA Machine Common Sense program and NSF Award 1922658 NRT-HDR: FUTURE Foundations, Translation, and Responsibility for Data Science.

\bibliography{baby_ssl}

\begin{thebibliography}{}

\bibitem[Ayzenberg and Lourenco, 2020]{ayzenberg2020}
Ayzenberg, V. and Lourenco, S. (2020).
\newblock Young children outperform feed-forward and recurrent neural networks on challenging object recognition tasks.
\newblock {\em Journal of Vision}, 20(11):310--310.

\bibitem[Bambach et~al., 2018]{bambach2018}
Bambach, S., Crandall, D., Smith, L., and Yu, C. (2018).
\newblock Toddler-inspired visual object learning.
\newblock {\em Advances in Neural Information Processing Systems}, 31.

\bibitem[Bomba and Siqueland, 1983]{bomba1983}
Bomba, P.~C. and Siqueland, E.~R. (1983).
\newblock The nature and structure of infant form categories.
\newblock {\em Journal of Experimental Child Psychology}, 35(2):294--328.

\bibitem[Caron et~al., 2021]{caron2021}
Caron, M., Touvron, H., Misra, I., J{\'e}gou, H., Mairal, J., Bojanowski, P., and Joulin, A. (2021).
\newblock Emerging properties in self-supervised vision transformers.
\newblock {\em arXiv preprint arXiv:2104.14294}.

\bibitem[Dosovitskiy et~al., 2020]{dosovitskiy2020}
Dosovitskiy, A., Beyer, L., Kolesnikov, A., Weissenborn, D., Zhai, X., Unterthiner, T., Dehghani, M., Minderer, M., Heigold, G., Gelly, S., et~al. (2020).
\newblock An image is worth 16x16 words: Transformers for image recognition at scale.
\newblock In {\em International Conference on Learning Representations}.

\bibitem[Elman et~al., 1996]{elman1996}
Elman, J.~L., Bates, E.~A., and Johnson, M.~H. (1996).
\newblock {\em Rethinking Innateness: A Connectionist Perspective on Development}, volume~10.
\newblock MIT Press.

\bibitem[Esser et~al., 2021]{esser2021}
Esser, P., Rombach, R., and Ommer, B. (2021).
\newblock Taming transformers for high-resolution image synthesis.
\newblock In {\em Proceedings of the IEEE/CVF Conference on Computer Vision and Pattern Recognition}, pages 12873--12883.

\bibitem[Geirhos et~al., 2020]{geirhos2020}
Geirhos, R., Jacobsen, J.-H., Michaelis, C., Zemel, R., Brendel, W., Bethge, M., and Wichmann, F.~A. (2020).
\newblock Shortcut learning in deep neural networks.
\newblock {\em Nature Machine Intelligence}, 2(11):665--673.

\bibitem[Geirhos et~al., 2021]{geirhos2021}
Geirhos, R., Narayanappa, K., Mitzkus, B., Thieringer, T., Bethge, M., Wichmann, F.~A., and Brendel, W. (2021).
\newblock Partial success in closing the gap between human and machine vision.
\newblock {\em Advances in Neural Information Processing Systems}, 34:23885--23899.

\bibitem[Grauman et~al., 2022]{grauman2022}
Grauman, K., Westbury, A., Byrne, E., Chavis, Z., Furnari, A., Girdhar, R., Hamburger, J., Jiang, H., Liu, M., Liu, X., et~al. (2022).
\newblock {Ego4D: Around the world in 3,000 hours of egocentric video}.
\newblock In {\em Proceedings of the IEEE/CVF Conference on Computer Vision and Pattern Recognition}, pages 18995--19012.

\bibitem[Gureckis and Markant, 2012]{gureckis2012}
Gureckis, T.~M. and Markant, D.~B. (2012).
\newblock Self-directed learning: {A} cognitive and computational perspective.
\newblock {\em {Perspectives on Psychological Science}}, 7(5):464--481.

\bibitem[He et~al., 2022]{he2022}
He, K., Chen, X., Xie, S., Li, Y., Doll{\'a}r, P., and Girshick, R. (2022).
\newblock Masked autoencoders are scalable vision learners.
\newblock In {\em Proceedings of the IEEE/CVF Conference on Computer Vision and Pattern Recognition}, pages 16000--16009.

\bibitem[Heusel et~al., 2017]{heusel2017}
Heusel, M., Ramsauer, H., Unterthiner, T., Nessler, B., and Hochreiter, S. (2017).
\newblock {GANs trained by a two time-scale update rule converge to a local Nash equilibrium}.
\newblock {\em Advances in Neural Information Processing Systems}, 30.

\bibitem[Huber et~al., 2022]{huber2022}
Huber, L.~S., Geirhos, R., and Wichmann, F.~A. (2022).
\newblock The developmental trajectory of object recognition robustness: children are like small adults but unlike big deep neural networks.
\newblock {\em arXiv preprint arXiv:2205.10144}.

\bibitem[Jabri et~al., 2020]{jabri2020}
Jabri, A., Owens, A., and Efros, A. (2020).
\newblock Space-time correspondence as a contrastive random walk.
\newblock {\em Advances in Neural Information Processing Systems}, 33:19545--19560.

\bibitem[Kellman and Spelke, 1983]{kellman1983}
Kellman, P.~J. and Spelke, E.~S. (1983).
\newblock Perception of partly occluded objects in infancy.
\newblock {\em Cognitive Psychology}, 15(4):483--524.

\bibitem[Kingma and Ba, 2014]{kingma2014}
Kingma, D.~P. and Ba, J. (2014).
\newblock {Adam: A method for stochastic optimization}.
\newblock {\em arXiv preprint arXiv:1412.6980}.

\bibitem[Konkle et~al., 2010]{konkle2010}
Konkle, T., Brady, T.~F., Alvarez, G.~A., and Oliva, A. (2010).
\newblock Conceptual distinctiveness supports detailed visual long-term memory for real-world objects.
\newblock {\em Journal of Experimental Psychology: General}, 139(3):558.

\bibitem[Lee et~al., 2021]{lee2021}
Lee, D., Gujarathi, P., and Wood, J.~N. (2021).
\newblock Controlled-rearing studies of newborn chicks and deep neural networks.
\newblock {\em arXiv preprint arXiv:2112.06106}.

\bibitem[Leibniz, 1704]{leibniz1704}
Leibniz, G.~W. (1704).
\newblock {\em Nouveaux essais sur l'entendement humain}.

\bibitem[Lillicrap et~al., 2020]{lillicrap2020}
Lillicrap, T.~P., Santoro, A., Marris, L., Akerman, C.~J., and Hinton, G. (2020).
\newblock Backpropagation and the brain.
\newblock {\em {Nature Reviews Neuroscience}}, 21(6):335--346.

\bibitem[Lin et~al., 2014]{lin2014}
Lin, T.-Y., Maire, M., Belongie, S., Hays, J., Perona, P., Ramanan, D., Doll{\'a}r, P., and Zitnick, C.~L. (2014).
\newblock {Microsoft COCO: Common objects in context}.
\newblock In {\em Computer Vision--ECCV 2014: 13th European Conference, Zurich, Switzerland, September 6-12, 2014, Proceedings, Part V 13}, pages 740--755. Springer.

\bibitem[Locatello et~al., 2020]{locatello2020}
Locatello, F., Weissenborn, D., Unterthiner, T., Mahendran, A., Heigold, G., Uszkoreit, J., Dosovitskiy, A., and Kipf, T. (2020).
\newblock Object-centric learning with slot attention.
\newblock {\em Advances in Neural Information Processing Systems}, 33:11525--11538.

\bibitem[Locke, 1690]{locke1690}
Locke, J. (1690).
\newblock {\em An Essay Concerning Human Understanding}.

\bibitem[Lomonaco and Maltoni, 2017]{lomonaco2017}
Lomonaco, V. and Maltoni, D. (2017).
\newblock {CORe50:} a new dataset and benchmark for continuous object recognition.
\newblock In {\em Conference on Robot Learning}, pages 17--26. PMLR.

\bibitem[Long et~al., 2022]{long2022}
Long, B., Goodin, S., Kachergis, G., Marchman, V.~A., Radwan, S., Sparks, R., Xiang, V., Zhuang, C., Hsu, O., Newman, B., et~al. (2022).
\newblock {The BabyView Camera:} designing a new head-mounted camera to capture children’s early social and visual environment.

\bibitem[Markman, 1990]{markman1990}
Markman, E.~M. (1990).
\newblock Constraints children place on word meanings.
\newblock {\em Cognitive Science}, 14(1):57--77.

\bibitem[Mehrer et~al., 2021]{mehrer2021}
Mehrer, J., Spoerer, C.~J., Jones, E.~C., Kriegeskorte, N., and Kietzmann, T.~C. (2021).
\newblock An ecologically motivated image dataset for deep learning yields better models of human vision.
\newblock {\em Proceedings of the National Academy of Sciences}, 118(8):e2011417118.

\bibitem[Merriman et~al., 1989]{merriman1989}
Merriman, W.~E., Bowman, L.~L., and MacWhinney, B. (1989).
\newblock The mutual exclusivity bias in children's word learning.
\newblock {\em Monographs of the Society for Research in Child Development}, pages i--129.

\bibitem[Moore et~al., 2022]{moore2022}
Moore, D.~S., Oakes, L.~M., Romero, V.~L., and McCrink, K.~C. (2022).
\newblock Leveraging developmental psychology to evaluate artificial intelligence.
\newblock In {\em {2022 IEEE International Conference on Development and Learning (ICDL)}}, pages 36--41. IEEE.

\bibitem[Murphy, 2004]{murphy2004}
Murphy, G. (2004).
\newblock {\em The Big Book of Concepts}.
\newblock MIT Press.

\bibitem[Orhan et~al., 2020]{orhan2020}
Orhan, E., Gupta, V., and Lake, B.~M. (2020).
\newblock Self-supervised learning through the eyes of a child.
\newblock {\em Advances in Neural Information Processing Systems}, 33.

\bibitem[Pont-Tuset et~al., 2017]{pont2017}
Pont-Tuset, J., Perazzi, F., Caelles, S., Arbel{\'a}ez, P., Sorkine-Hornung, A., and Van~Gool, L. (2017).
\newblock The 2017 {DAVIS} challenge on video object segmentation.
\newblock {\em arXiv preprint arXiv:1704.00675}.

\bibitem[Radford et~al., 2019]{radford2019}
Radford, A., Wu, J., Child, R., Luan, D., Amodei, D., Sutskever, I., et~al. (2019).
\newblock Language models are unsupervised multitask learners.
\newblock {\em OpenAI preprint}.

\bibitem[Russakovsky et~al., 2015]{russakovsky2015}
Russakovsky, O., Deng, J., Su, H., Krause, J., Satheesh, S., Ma, S., Huang, Z., Karpathy, A., Khosla, A., Bernstein, M., et~al. (2015).
\newblock {ImageNet} large scale visual recognition challenge.
\newblock {\em {International Journal of Computer Vision}}, 115(3):211--252.

\bibitem[Smaira et~al., 2020]{smaira2020}
Smaira, L., Carreira, J., Noland, E., Clancy, E., Wu, A., and Zisserman, A. (2020).
\newblock A short note on the {Kinetics-700-2020} human action dataset.
\newblock {\em arXiv preprint arXiv:2010.10864}.

\bibitem[Smith and Slone, 2017]{smith2017}
Smith, L.~B. and Slone, L.~K. (2017).
\newblock A developmental approach to machine learning?
\newblock {\em Frontiers in Psychology}, page 2124.

\bibitem[Spelke, 1994]{spelke1994}
Spelke, E. (1994).
\newblock Initial knowledge: Six suggestions.
\newblock {\em Cognition}, 50(1-3):431--445.

\bibitem[Spelke et~al., 1992]{spelke1992}
Spelke, E.~S., Breinlinger, K., Macomber, J., and Jacobson, K. (1992).
\newblock Origins of knowledge.
\newblock {\em Psychological Review}, 99(4):605.

\bibitem[Sullivan et~al., 2022]{sullivan2022}
Sullivan, J., Mei, M., Perfors, A., Wojcik, E., and Frank, M.~C. (2022).
\newblock {SAYCam: A large, longitudinal audiovisual dataset recorded from the infant’s perspective}.
\newblock {\em Open Mind}, 5:20--29.

\bibitem[Van~der Maaten and Hinton, 2008]{vandermaaten2008}
Van~der Maaten, L. and Hinton, G. (2008).
\newblock {Visualizing data using t-SNE}.
\newblock {\em {Journal of Machine Learning Research}}, 9(11).

\bibitem[Xie et~al., 2017]{xie2017}
Xie, S., Girshick, R., Doll{\'a}r, P., Tu, Z., and He, K. (2017).
\newblock Aggregated residual transformations for deep neural networks.
\newblock In {\em Proceedings of the IEEE Conference on Computer Vision and Pattern Recognition}, pages 1492--1500.

\bibitem[Zaadnoordijk et~al., 2022]{zaadnoordijk2022}
Zaadnoordijk, L., Besold, T.~R., and Cusack, R. (2022).
\newblock Lessons from infant learning for unsupervised machine learning.
\newblock {\em Nature Machine Intelligence}, 4(6):510--520.

\bibitem[Zhou et~al., 2016]{zhou2016}
Zhou, B., Khosla, A., Lapedriza, A., Oliva, A., and Torralba, A. (2016).
\newblock Learning deep features for discriminative localization.
\newblock In {\em Proceedings of the IEEE Conference on Computer Vision and Pattern Recognition}, pages 2921--2929.

\bibitem[Zhou et~al., 2017]{zhou2017}
Zhou, B., Lapedriza, A., Khosla, A., Oliva, A., and Torralba, A. (2017).
\newblock Places: A 10 million image database for scene recognition.
\newblock {\em IEEE Transactions on Pattern Analysis and Machine Intelligence}.

\bibitem[Zhou et~al., 2022]{zhou2022}
Zhou, P., Zhou, Y., Si, C., Yu, W., Ng, T.~K., and Yan, S. (2022).
\newblock Mugs: A multi-granular self-supervised learning framework.
\newblock {\em arXiv preprint arXiv:2203.14415}.

\bibitem[Zhuang et~al., 2022]{zhuang2022}
Zhuang, C., Xiang, V., Bai, Y., Jia, X., Turk-Browne, N., Norman, K., DiCarlo, J.~J., and Yamins, D.~L. (2022).
\newblock How well do unsupervised learning algorithms model human real-time and life-long learning?
\newblock In {\em Thirty-sixth Conference on Neural Information Processing Systems Datasets and Benchmarks Track}.

\bibitem[Zhuang et~al., 2021]{zhuang2021}
Zhuang, C., Yan, S., Nayebi, A., Schrimpf, M., Frank, M.~C., DiCarlo, J.~J., and Yamins, D.~L. (2021).
\newblock Unsupervised neural network models of the ventral visual stream.
\newblock {\em Proceedings of the National Academy of Sciences}, 118(3):e2014196118.

\end{thebibliography}
\bibliographystyle{apalike}

\section*{Appendix}

\subsubsection*{Evaluation tasks for the embedding models}
Here, we describe the tasks used for evaluating the embedding models:

\textbf{Labeled S:} Labeled S contains a total of $\sim$58K manually labeled frames from child S in SAYCam \citep{orhan2020}. We use the temporally 10$\times$ subsampled version of this dataset (0.1 frames/second) containing $\sim$5.8K images from 26 different classes. Temporal subsampling reduces the temporal correlations in the dataset and makes the classification task more challenging. We then randomly split the data in half, use the first half for training and the second half for evaluation. This is our only within-domain evaluation task for models trained on SAYCam, specifically for models trained on data from child S.

\textbf{Konkle objects:} This is a public dataset available from \href{https://bradylab.ucsd.edu/stimuli.html}{this address}. The images in this dataset depict common everyday objects in isolation against a uniform white background \citep{konkle2010}. We only use a subset of the categories from the dataset that contain a sufficiently large number of exemplars, \textit{i.e.} 16 or 17 exemplars. This subset contains 4040 images from 240 different object categories. We split the data in half, use the first half for training and the second half for evaluation. 

\textbf{CORe50:} This is a public dataset available from \href{https://vlomonaco.github.io/core50}{this address}. The dataset contains 50 different everyday objects undergoing various continuous transformations (complex combinations of 3D rotations and translations) against a variety of backgrounds \citep{lomonaco2017}. The dataset is originally in video format, but we sample the videos at 5 frames/second to make an image dataset out of it. Each object is shot against the same set of 11 unique backgrounds. We use 6 of these backgrounds for training and the remaining 5 backgrounds for evaluation (90K images in total for training, 75K images for evaluation). This task thus tests whether (i) a model can ignore the background and primarily respond to the foreground object instead and (ii) generalize over continuous transformations. Note that a model primarily responding to the background would perform at near chance levels (2\% top-1 accuracy) in this task, since the background does not have any predictive value for the object identity.   

\textbf{ImageNet:} ImageNet (ILSVRC-2012) is a large and diverse dataset of high-quality images from the internet \citep{russakovsky2015} and is a very popular benchmark for real-world visual object recognition. The dataset is publicly available from \href{https://www.image-net.org}{this address}. We use the standard training-validation split for this dataset, containing $\sim$1.28M training images and 50K validation images from 1000 semantic classes.

\textbf{ImageNet OOD:} To evaluate the robustness, or out-of-distribution (OOD) generalization capabilities, of the trained models, we also consider out-of-distribution versions of the ImageNet benchmark \citep{geirhos2020,geirhos2021}. The ImageNet OOD benchmark contains 17 different out-of-distribution versions of ImageNet generated by applying various transformations to images from the ImageNet validation set. These include transformations such as taking the silhouettes of the objects in the image, stylizing the image, adding different types of noise to the image, changing the colors in the image, etc. For evaluation, we use the OOD accuracy metric, which is just the mean top-1 accuracy over all 17 out-of-distribution datasets \citep{geirhos2021}. This evaluation dataset is publicly available from \href{https://github.com/bethgelab/model-vs-human}{this address}.

\textbf{Ecoset:} Ecoset can be thought of as an ecologically more realistic version of ImageNet containing images from 565 basic-level categories only, selected for their concreteness and frequency of usage in language \citep{mehrer2021}. The dataset comes with a standard training-validation split containing $\sim$1.44M training images and 28250 validation images, which we use for training and evaluation respectively. The dataset is publicly available from \href{https://huggingface.co/datasets/kietzmannlab/ecoset}{this address}.  

\textbf{Places365:} Because the SAYCam dataset contains examples of various scene categories (\textit{living room, dining room, kitchen, bathroom, playground, beach, street, porch, etc.}) in addition to object categories, we are interested in evaluating the capacity of SAYCam-trained models to recognize places as well as objects. For this purpose, we use the Places365 dataset \citep{zhou2017}. Places365 contains $\sim$1.8M training images and 36500 validation images from 365 different place categories. The dataset is publicly available from  \href{http://places2.csail.mit.edu}{this address}.

\textbf{DAVIS-2017:} A good visual representation is ideally a general-purpose representation that can be used profitably not just in visual recognition tasks, but in a broader range of downstream tasks. For this reason, we also evaluate the SAYCam-learned representations in two dense prediction tasks. DAVIS-2017 is a video object segmentation task where the model is given a ground-truth segmentation mask for the initial frame of a short video clip and is expected to predict the segmentation masks for the following frames in the video \citep{pont2017}. In common evaluation protocols used for this task, the predicted segmentation masks for the non-initial frames are computed with a non-parametric message passing type algorithm that uses the representations of the frames and the predicted segmentation masks for nearby frames. This task essentially evaluates how robust the model's representations of the objects in the video clip are to spatio-temporal transformations that take place in the clip: more robust representations are expected to propagate the initial ground-truth segmentation masks better. The evaluation set consists of 30 video clips, each containing $\sim$67 frames and $\sim$2 objects on average. The data are publicly available from \href{https://davischallenge.org/davis2017/code.html}{this address}.

\textbf{COCO:} We also evaluate our models on the semantic segmentation component of the COCO benchmark \citep{lin2014}. COCO is publicly available to download from \href{https://cocodataset.org}{this address}. Recall that in semantic segmentation the goal is to label each pixel of the image with the semantic category label of the object (or ``stuff'') occupying that pixel. We use a subset of COCO that contains the 21 categories present in the Pascal VOC dataset. This subset has $\sim$92.5K training images and 5K validation images in total.  

For all evaluation tasks except DAVIS-2017 (including the COCO semantic segmentation task), we use linear readouts trained on top of frozen features, also known as a linear probe. For DAVIS-2017, as mentioned above, we use a standard non-parametric label propagation algorithm to predict the segmentation masks \citep{jabri2020}. We use standard evaluation metrics for all our evaluation tasks: top-1 accuracy for the classification tasks, mean intersection over union (IoU) for the COCO semantic segmentation task, and the mean region and contour similarity ($\mathcal{J\&F}$) for DAVIS-2017.

\subsubsection*{SSL algorithms for the embedding models}
Here, we describe each of the three SSL algorithms we used for training our embedding models. These algorithms represent a range of different modern approaches to self-supervised representation learning from static images or frames.

\textbf{DINO:} DINO is a self-distillation type representation learning algorithm \citep{caron2021}, where a teacher model and a student model iteratively improve each other. During training, the teacher and the student receive different copies of the same image, transformed in various ways with a set of data augmentation methods, and the objective of the algorithm is to push the representations of these copies towards each other, because they share the same semantic content. The data augmentation methods used in DINO are color jitter, random resized crops, horizontal flips, gray-scaling, Gaussian blur, and solarization.

\textbf{Mugs:} Mugs is a hybrid SSL algorithm combining ideas from self-distillation and contrastive learning to learn multi-granular visual representations \citep{zhou2022}. Mugs uses the same set of data augmentations as DINO.

\textbf{MAE:} Masked autoencoders (MAEs) use reconstruction of masked image patches as the self-supervised learning objective \citep{he2022}. By learning to predict masked patches from visible patches, the algorithm expects to learn higher level, semantically useful regularities in visual scenes (\textit{e.g.} learning that the face, the legs, and the tail of a dog often appear in a particular configuration). MAEs use a much lighter data augmentation pipeline than other algorithms, requiring only random resized crops and horizontal flips. As recommended \citep{he2022}, we use a large masking ratio of 75\% during training, \textit{i.e.}~75\% of the image patches are randomly masked out.

We generally use the default hyperparameter choices and training configurations recommended for these algorithms in the original papers, with minor modifications. We use the same data augmentation pipeline for every model trained with a given algorithm. Further details can be found in the corresponding training codes that can be accessed from our main public repository.

\subsubsection*{Reference datasets for the embedding models}
Kinetics-700 consists of short YouTube clips of people performing various actions, representing 700 different action categories \citep{smaira2020}. Kinetics-700 is publicly available for download from \href{https://www.deepmind.com/open-source/kinetics}{this address}. The video clips in Kinetics-700 are typically shorter than 10 seconds, hence the dataset overall is expected to be much more diverse in style and content and temporally much less correlated than SAYCam. Ego4D, on the other hand, has more similar temporal characteristics to SAYCam: the videos are temporally extended, continuous, egocentric headcam recordings, with recording sessions lasting tens of minutes on average \citep{grauman2022}. The main differences from SAYCam are (i) the videos are taken from the perspective of adult camera wearers, not from the perspective of young children, and (ii) the recordings are made by many more individuals than the SAYCam recordings: in Ego4D, each individual contributes $\sim$4 hours of recording on average, so a 200-hour long subset of the dataset would be expected to contain recordings from roughly 50 different camera wearers, in contrast to a single child in SAYCam. Ego4D is publicly available from \href{https://ego4d-data.org/}{this address} (after signing a license agreement). We use 200-hour long subsets of these datasets, because 200 hours is roughly equal to the total length of the video data we have available from one of the children in SAYCam, namely S. To obtain these 200-hour long subsets, we use the first 128 clips from each class in Kinetics-700 and select a continuous chunk of videos from Ego4D with a random starting point until the total length of the videos in the selection roughly equals 200 hours.

\subsubsection*{Training details for the embedding models}
We train each model for four days on four A100 GPUs (with 80 GB GPU memory) using data parallelism (the ViT-B/14 DINO model trained on all of ImageNet was trained for four additional days to make sure it was not under-trained). We use the Adam optimizer to train all models \citep{kingma2014}. In each experiment, we use either a batch size of 512 or the largest batch size we could fit on four GPUs (in cases where we could not fit a total batch size of 512 on the GPUs). Batch sizes and learning rates thus vary across experiments. Inspection of the training losses confirms that they all saturate, hence under-training is unlikely for any of our pretraining runs (all training logs are made available in our public repository).~Table~\ref{embedding_models_table} presents a concise list of all embedding models trained for this work.

\subsubsection*{Class activation maps (CAMs)}
In visualizing the class activation maps (CAMs) shown in Figure~\ref{atts_cams_fig}a, we first normalize the linearly combined and upsampled feature map to have zero mean and unit variance, where the mean and variance are estimated over a batch of images from the same class, pass the normalized map through a pointwise sigmoid nonlinearity, and then scale it by 255 so that the values in the final map are between 0 and 255 (or in \texttt{torch} notation: \texttt{m = 255 * torch.sigmoid((m-torch.mean(m))/torch.std(m))}. We then alpha-blend this activation map with the original image using a blending coefficient of 0.8 for the map and 0.2 for the image.

\subsubsection*{Additional details about the generative models}
We train customized VQGAN models using the \href{https://github.com/CompVis/taming-transformers}{Taming Transformers} repository made available by the authors of VQGAN \cite{esser2021}. For the GPT model, we use a standard 730M-parameter GPT model that is similar to OpenAI's \texttt{gpt2-large} model \citep{radford2019}. Using the same architecture, we also train reference VQGAN-GPT models on ImageNet, using either 100\%, 10\%, or 1\% of the training set, as described previously.

For the VQGAN component of the generative models for SAYCam, we use a codebook with a vocabulary size of 8192 and a spatial resolution of $32\times 32$ (thus each frame is encoded as a $32\times 32$ grid of integers, where the integers take values between 1 and 8192). For the encoded SAYCam frames, the spatial resolution of $32\times 32$ corresponds to a sequence length of 1024 tokens. Due to computational constraints, the VQGAN models for ImageNet use a spatial resolution of $16\times 16$ and a codebook with a dictionary size of 16384. To train the VQGAN component of the generative model, we use the \href{https://github.com/CompVis/taming-transformers}{Taming Transformers} repository made available by the authors of VQGAN \cite{esser2021} (model configuration files are available from our public repository). The GPT component of the generative models has 36 layers, 20 attention heads, and an embedding dimensionality of 1280 in all cases (the model configuration is equivalent to OpenAI's \texttt{gpt2-large} model). We generate the model completions through exact sampling, with the softmax temperature set to $T=1.0$. 

\subsubsection*{Training and evaluation details for the generative models}
SAYCam-trained GPT models were trained for four days on 16 A100 GPUs with a batch size of 96 (the model trained on the combined data from SAYCam was trained for four additional days to make sure it was not under-trained). The training logs (all made available from our public repository) confirm that under-training is not a serious concern for any of our models. The ImageNet-trained models were trained on 8 A100 GPUs with a total batch size of 256 (the model trained on 100\% of ImageNet was trained for 6 days, whereas the models trained on 10\% and 1\% of ImageNet were trained for 2 days only due to the more limited size of the training data in these cases). All models were trained with the Adam algorithm. Table~\ref{generative_models_table} presents a concise list of all generative models trained for this work.

We measure the overall quality of the completions with the Fr\'echet Inception distance (FID) between the model generated samples and the ground-truth images \citep{heusel2017}. We use two qualitatively very different datasets for evaluation: \textit{Labeled S} and \textit{Konkle objects} (both described in more detail above).

\subsubsection*{Quantitative evaluation of the generative models}
We use three different image completion tasks to quantitatively evaluate the generative models: Labeled S, Konkle (\textit{iid}), and Konkle (\textit{ood}). In Labeled S, we use images from the validation split of the \textit{Labeled S} dataset described in the main text for the image completion task. In Konkle (\textit{iid}), we randomly split the \textit{Konkle objects} dataset in half, use the first half for training or finetuning the generative models, and use the other half for the image completion task. In Konkle (\textit{ood}), we split the \textit{Konkle objects} dataset into non-overlapping vehicle and non-vehicle categories, use the non-vehicle categories for training or finetuning the generative models, and use the vehicle categories (144 images in total) for the image completion task. Since this is an out-of-distribution generalization task, it is expected to be more challenging than the \textit{iid} condition. The results are presented in Table~\ref{fid_scores_table} below showing the FID scores of different models in each image completion task. Finetuning the SAYCam-trained models with a few thousand images from the \textit{Konkle objects} dataset improves their generation quality both in \textit{iid} and \textit{ood} conditions.

\subsubsection*{Data availability}
With the exception of SAYCam, all data used in this study are publicly available. Instructions for accessing the public datasets are detailed above. The SAYCam dataset can be accessed by authorized users with an institutional affiliation from the following Databrary repository: \url{http://doi.org/10.17910/b7.564}. The \textit{Labeled S} evaluation dataset, which is a subset of SAYCam, is also available from the same repository under the session name \textit{Labeled S}.

\subsubsection*{Code availability}
All our pretrained models (over 70 different models) as well as a variety of tools to use and analyze them are available from the following public repository: \url{https://github.com/eminorhan/silicon-menagerie}. The code used for training and evaluating the models is also publicly available from the same repository.

\begin{figure}
    \centering
    \includegraphics[width=0.93\textwidth]{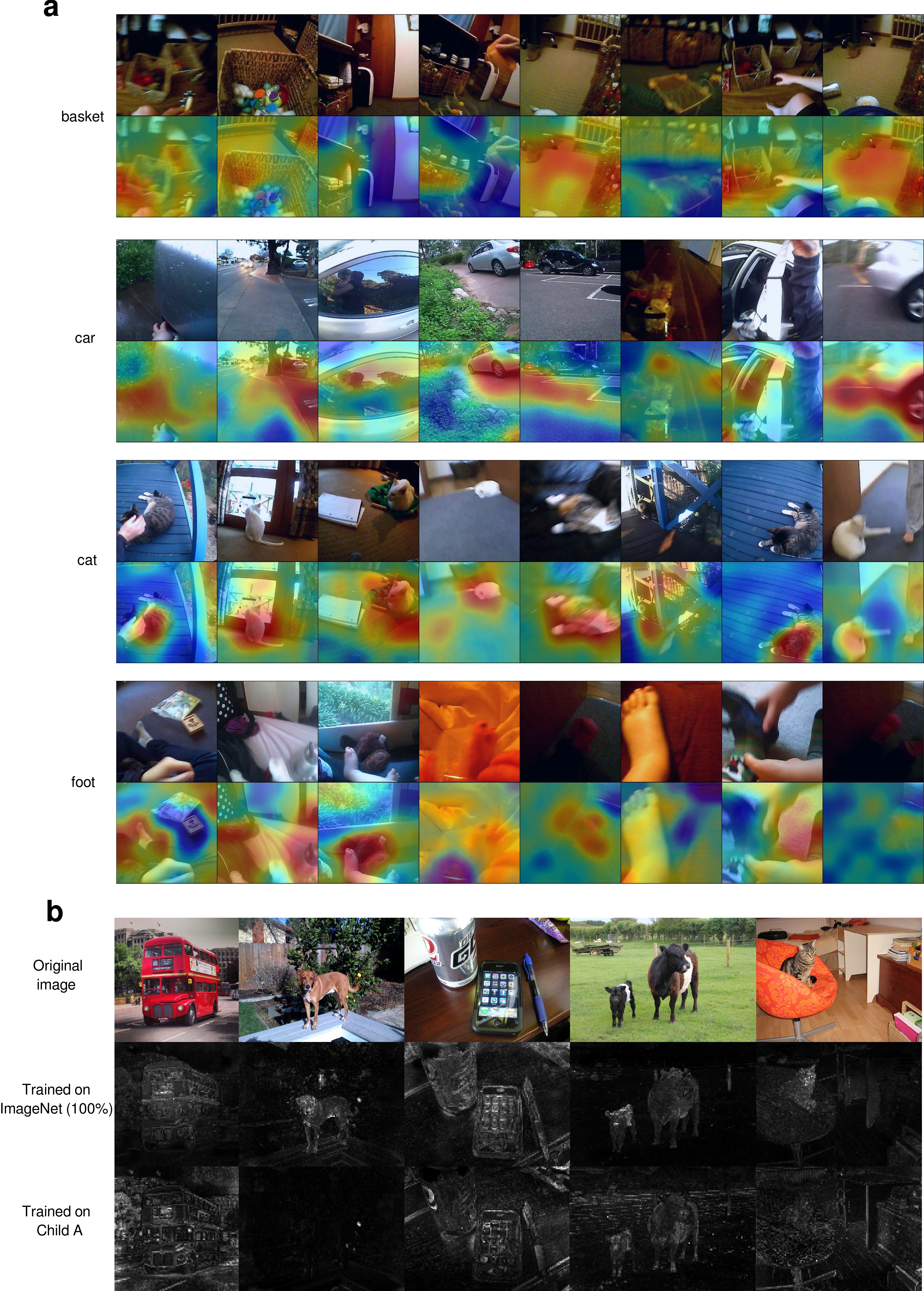}
    \caption{(\textbf{a}) Class activation maps (CAMs) for a ResNeXt-50 model trained with DINO on the headcam data from child A. (\textbf{b}) Mean attention maps (averaged over all attention heads) for a ViT-B/14 model trained with DINO on all of ImageNet or on the headcam data from child A, respectively.}
    \label{atts_a_fig}
\end{figure}

\begin{figure}
    \centering
    \includegraphics[width=0.93\textwidth]{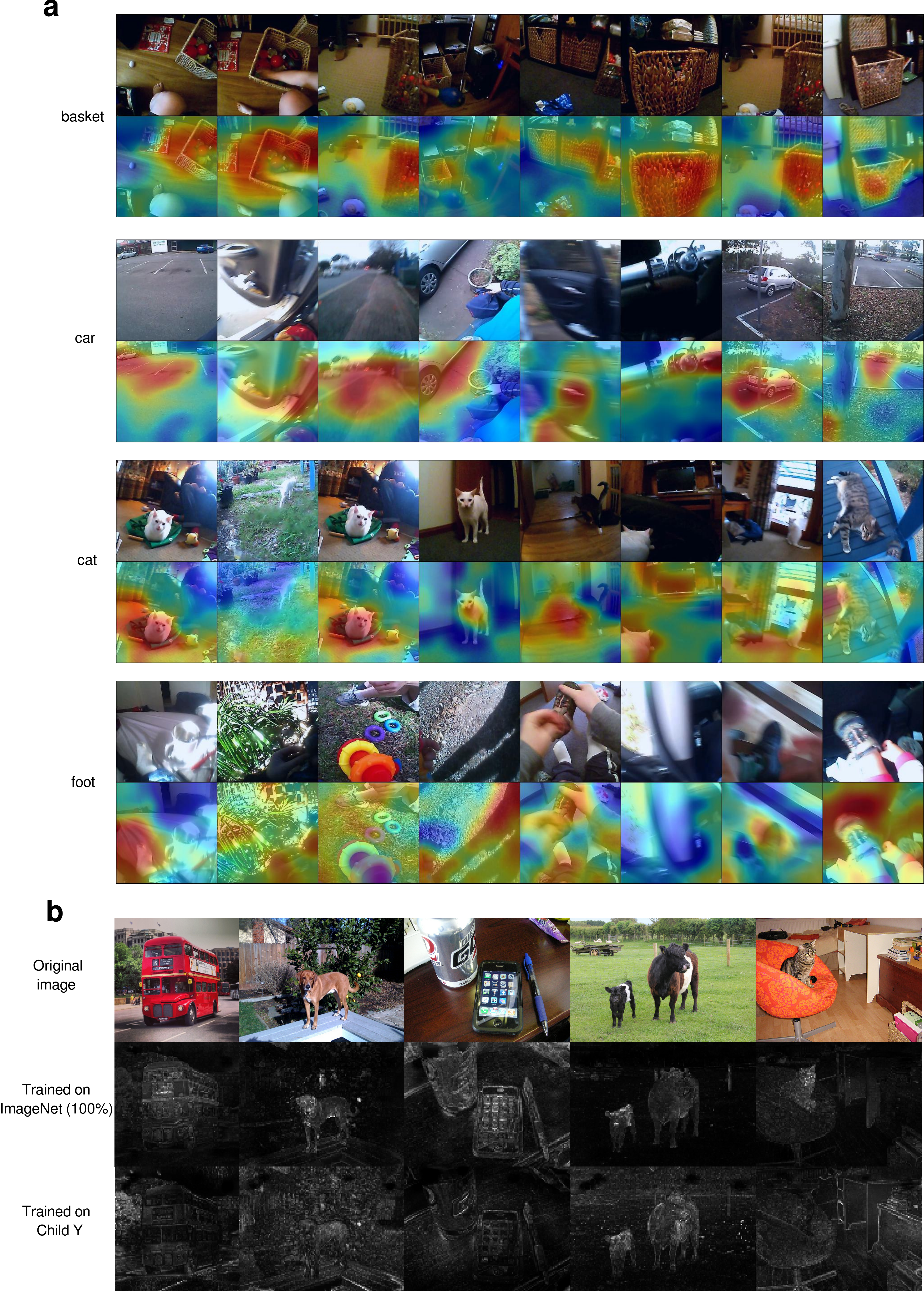}
    \caption{(\textbf{a}) Class activation maps (CAMs) for a ResNeXt-50 model trained with DINO on the headcam data from child Y. (\textbf{b}) Mean attention maps (averaged over all attention heads) for a ViT-B/14 model trained with DINO on all of ImageNet or on the headcam data from child Y, respectively.}
    \label{atts_y_fig}
\end{figure}

\begin{figure}
    \centering
    \includegraphics[width=1.0\textwidth]{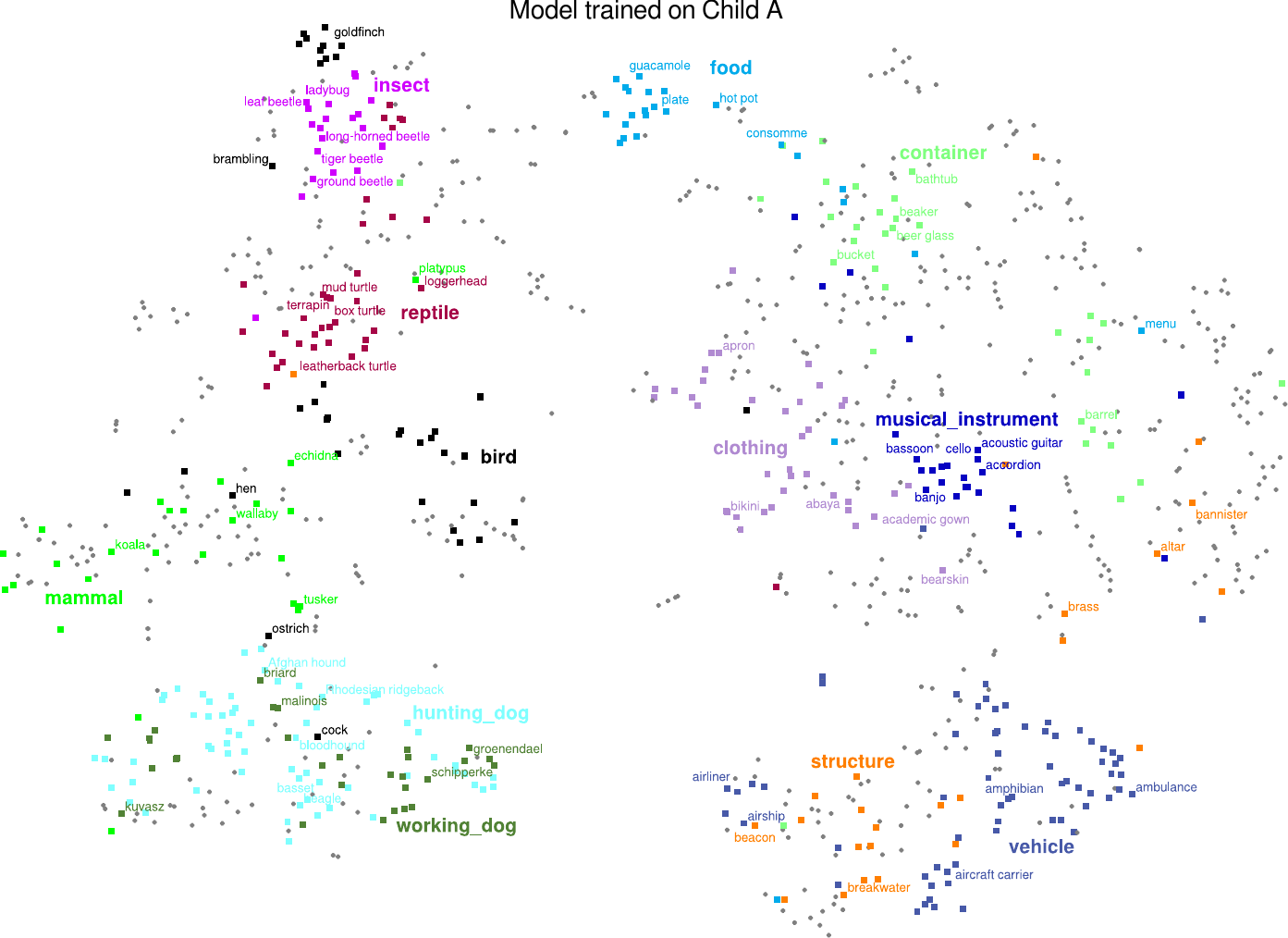}
    \caption{t-SNE visualization of the 1000 ImageNet classes for a ViT-B/14 model trained with DINO on the headcam data from child A.}
    \label{tsne_a_fig}
\end{figure}

\begin{figure}
    \centering
    \includegraphics[width=1.0\textwidth]{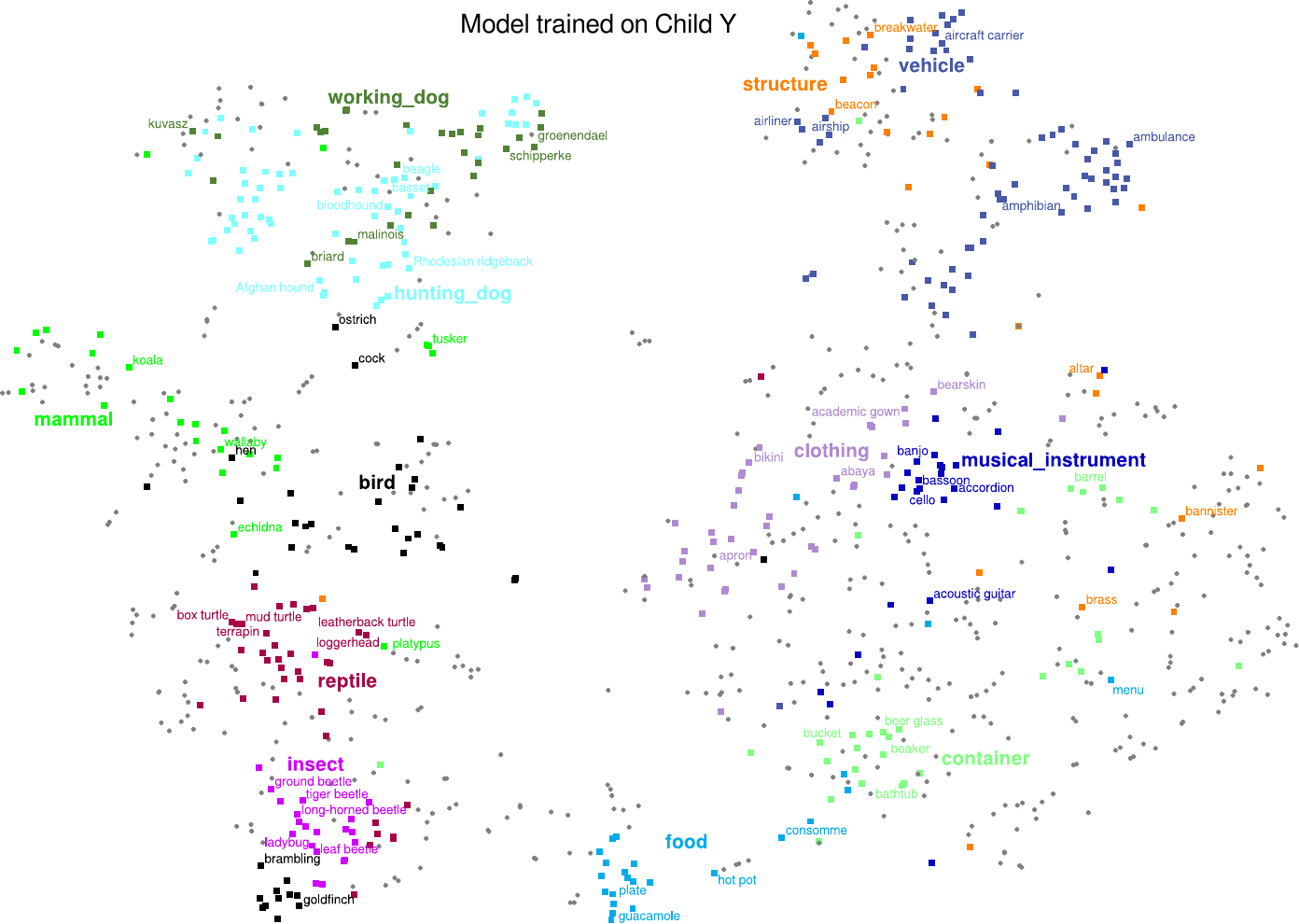}
    \caption{t-SNE visualization of the 1000 ImageNet classes for a ViT-B/14 model trained with DINO on the headcam data from child Y.}
    \label{tsne_y_fig}
\end{figure}

\begin{figure}
    \centering
    \includegraphics[width=1.0\textwidth]{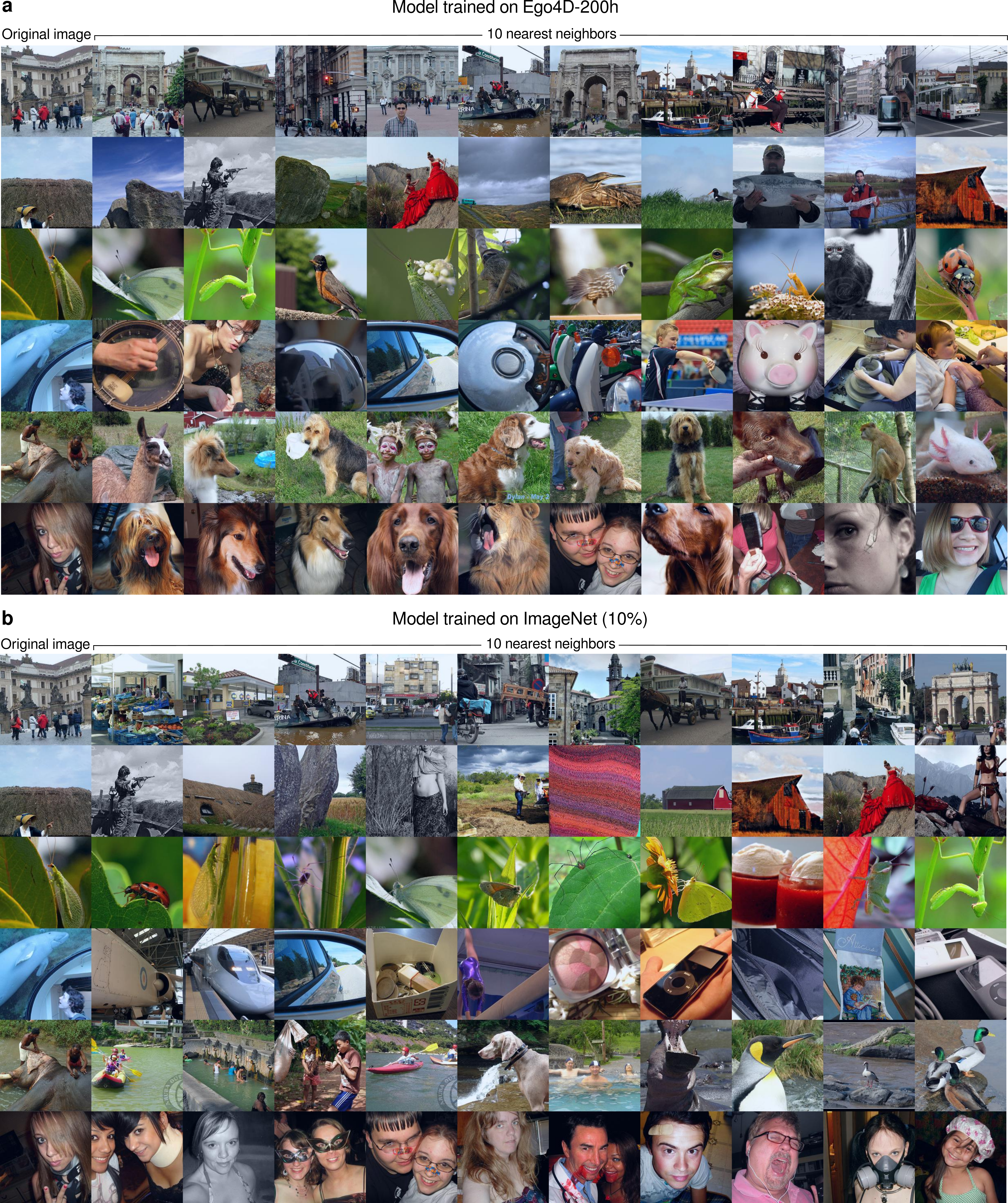}
    \caption{Nearest neighbors in the embedding space of ViT-B/14 models trained with DINO on the Ego4D-200h dataset (\textbf{a}) and on 10\% of ImageNet (\textbf{b}).}
    \label{nearest_neighbors_ego4d_imagenet10_fig}
\end{figure}

\vspace{1cm}

\begin{table}[h]
    \centering
\begin{tabular}{ccccccc}
\multirow{2}{*}{\textbf{Algorithms}}       & \multirow{2}{*}{\textbf{Data}}        & \multicolumn{5}{c}{\textbf{Models}}                    \\ 
                                           &                                       & ResNeXt-50 & ViT-B/14 & ViT-L/16 & ViT-B/16 & ViT-S/16 \\ \hline
\multicolumn{1}{c|}{\multirow{9}{*}{DINO}} & \multicolumn{1}{c|}{SAY}              & \checkmark & \checkmark & \checkmark & \checkmark & \checkmark \\
\multicolumn{1}{c|}{}                      & \multicolumn{1}{c|}{S}                & \checkmark & \checkmark & \checkmark & \checkmark & \checkmark \\
\multicolumn{1}{c|}{}                      & \multicolumn{1}{c|}{A}                & \checkmark & \checkmark & \checkmark & \checkmark & \checkmark \\
\multicolumn{1}{c|}{}                      & \multicolumn{1}{c|}{Y}                & \checkmark & \checkmark & \checkmark & \checkmark & \checkmark \\
\multicolumn{1}{c|}{}                      & \multicolumn{1}{c|}{Ego4D-200h}       &            & \checkmark &          &          &          \\
\multicolumn{1}{c|}{}                      & \multicolumn{1}{c|}{Kinetics-200h}    &            & \checkmark &          &          &          \\
\multicolumn{1}{c|}{}                      & \multicolumn{1}{c|}{ImageNet (100\%)} &            & \checkmark &          &          &          \\
\multicolumn{1}{c|}{}                      & \multicolumn{1}{c|}{ImageNet (10\%)}  &            & \checkmark &          &          &          \\
\multicolumn{1}{c|}{}                      & \multicolumn{1}{c|}{ImageNet (1\%)}   &            & \checkmark &          &          &          \\ \hline
\multicolumn{1}{c|}{\multirow{4}{*}{Mugs}} & \multicolumn{1}{c|}{SAY}              &            &          & \checkmark & \checkmark & \checkmark \\
\multicolumn{1}{c|}{}                      & \multicolumn{1}{c|}{S}                &            &          & \checkmark & \checkmark & \checkmark \\
\multicolumn{1}{c|}{}                      & \multicolumn{1}{c|}{A}                &            &          & \checkmark & \checkmark & \checkmark \\
\multicolumn{1}{c|}{}                      & \multicolumn{1}{c|}{Y}                &            &          & \checkmark & \checkmark & \checkmark \\ \hline
\multicolumn{1}{c|}{\multirow{4}{*}{MAE}}  & \multicolumn{1}{c|}{SAY}              &            &          & \checkmark & \checkmark & \checkmark \\
\multicolumn{1}{c|}{}                      & \multicolumn{1}{c|}{S}                &            &          & \checkmark & \checkmark & \checkmark \\
\multicolumn{1}{c|}{}                      & \multicolumn{1}{c|}{A}                &            &          & \checkmark & \checkmark & \checkmark \\
\multicolumn{1}{c|}{}                      & \multicolumn{1}{c|}{Y}                &            &          & \checkmark & \checkmark & \checkmark        
\end{tabular}
    \caption{List of all trained embedding models (49 models in total).~The trained algorithm$\times$data$\times$model combinations are indicated by check marks.}
    \label{embedding_models_table}
\end{table}

\begin{table}[h]
    \centering
\begin{tabular}{cccc}
\multirow{2}{*}{\textbf{Pretraining data}}         & \multicolumn{3}{c}{\textbf{Finetuning data}} \\
                                                   & Konkle (\textit{iid}) & Konkle (\textit{non-vehicle}) & None \\ \hline
\multicolumn{1}{c|}{SAY}              & \checkmark & \checkmark & \checkmark \\
\multicolumn{1}{c|}{S}                & \checkmark & \checkmark & \checkmark \\
\multicolumn{1}{c|}{A}                & \checkmark & \checkmark & \checkmark \\
\multicolumn{1}{c|}{Y}                & \checkmark & \checkmark & \checkmark \\
\multicolumn{1}{c|}{ImageNet (100\%)} & \checkmark & \checkmark & \checkmark \\
\multicolumn{1}{c|}{ImageNet (10\%)}  & \checkmark & \checkmark & \checkmark \\
\multicolumn{1}{c|}{ImageNet (1\%)}   & \checkmark & \checkmark & \checkmark \\
\multicolumn{1}{c|}{None}             & \checkmark & \checkmark &  \\

\end{tabular}
    \caption{List of all trained generative models (23 models in total).~The trained pretraining$\times$finetuning data combinations are indicated by check marks. `None' means pretraining (or finetuning) was not applied.}
    \label{generative_models_table}
\end{table}

\begin{table}[h]
    \centering
\begin{tabular}{cccc}
\multirow{2}{*}{\textbf{Models}}         & \multicolumn{3}{c}{\textbf{Tasks}} \\
                                      & Labeled S & Konkle (\textit{iid}) & Konkle (\textit{ood}) \\ \hline
\multicolumn{1}{c|}{SAY}              & 44.5 & 45.4 & 163.6 \\
\multicolumn{1}{c|}{S}                & 45.9 & 55.1 & 166.4 \\
\multicolumn{1}{c|}{A}                & 60.7 & 63.1 & 189.3 \\
\multicolumn{1}{c|}{Y}                & 60.4 & 63.6 & 176.8 \\
\multicolumn{1}{c|}{ImageNet (100\%)} & 62.8 & 24.0 & 89.6 \\
\multicolumn{1}{c|}{ImageNet (10\%)}  & 65.7 & 31.3 & 115.4 \\
\multicolumn{1}{c|}{ImageNet (1\%)}   & 81.7 & 43.5 & 146.0 \\ \hline
\multicolumn{1}{c|}{Konkle \textit{iid}}   & -- & 33.1 & -- \\
\multicolumn{1}{c|}{SAY+Konkle \textit{iid}}   & -- & 30.0 & -- \\
\multicolumn{1}{c|}{S+Konkle \textit{iid}}   & -- & 31.5 & -- \\
\multicolumn{1}{c|}{A+Konkle \textit{iid}}   & -- & 35.4 & -- \\
\multicolumn{1}{c|}{Y+Konkle \textit{iid}}   & -- & 31.9 & -- \\
\multicolumn{1}{c|}{ImageNet (100\%)+Konkle \textit{iid}}   & -- & 23.5 & -- \\
\multicolumn{1}{c|}{ImageNet (10\%)+Konkle \textit{iid}}   & -- & 28.4 & -- \\
\multicolumn{1}{c|}{ImageNet (1\%)+Konkle \textit{iid}}   & -- & 31.2 & -- \\ \hline
\multicolumn{1}{c|}{Konkle non-vehicle}   & -- & -- & 137.9 \\ 
\multicolumn{1}{c|}{SAY+Konkle non-vehicle}   & -- & -- & 136.9 \\
\multicolumn{1}{c|}{S+Konkle non-vehicle}   & -- & -- & 136.1 \\
\multicolumn{1}{c|}{A+Konkle non-vehicle}   & -- & -- & 138.8 \\
\multicolumn{1}{c|}{Y+Konkle non-vehicle}   & -- & -- & 143.0 \\
\multicolumn{1}{c|}{ImageNet (100\%)+Konkle non-vehicle}   & -- & -- & 124.1 \\
\multicolumn{1}{c|}{ImageNet (10\%)+Konkle non-vehicle}   & -- & -- & 141.3 \\
\multicolumn{1}{c|}{ImageNet (1\%)+Konkle non-vehicle}   & -- & -- & 143.1 \\

\end{tabular}
    \caption{FID scores in conditional generation tasks. Lower scores indicate better results (generated samples are more similar to ground-truth images). Rows correspond to different models, columns correspond to different tasks. Models are identified by the data they are trained and finetuned with. Model names in the format \texttt{`x+y'} mean the model was first trained on \texttt{x} and then finetuned on \texttt{y}.}
    \label{fid_scores_table}
\end{table}


\end{document}